%% file: main.tex
\documentclass[10pt,twocolumn,letterpaper]{article}
\usepackage[pagenumbers]{cvpr}      % To produce the REVIEW version

\usepackage{graphicx}
\usepackage{amsmath}
\usepackage{amssymb}
\usepackage{booktabs}

\usepackage[pagebackref,breaklinks,colorlinks]{hyperref}

\usepackage[capitalize]{cleveref}
\crefname{section}{Sec.}{Secs.}
\Crefname{section}{Section}{Sections}
\Crefname{table}{Table}{Tables}
\crefname{table}{Tab.}{Tabs.}

\def\cvprPaperID{6579} % *** Enter the CVPR Paper ID here
\def\confName{CVPR}
\def\confYear{2023}

\newcommand{\model}{\mathcal{M}}
\newcommand{\classes}{\mathcal{C}}
\newcommand{\fe}{\mathcal{F}}
\newcommand{\classifier}{\mathcal{G}}
\newcommand{\data}{\mathcal{D}}
\newcommand{\acc}[2]{$\text{Acc}({#1}, {#2})$}

\newcommand{\arxiv}{\textcolor{black}}

\begin{document}

\title{On the Stability-Plasticity Dilemma of Class-Incremental Learning}

\author{Dongwan Kim$^1$ \qquad\qquad Bohyung Han$^{1, 2}$ \\
Computer Vision Laboratory, ECE$^1$ \& IPAI$^{2}$, Seoul National University\\
%Institution1 address\\
{\tt\small \{dongwan123, bhhan\}@snu.ac.kr}
}
\maketitle

\begin{abstract}
    A primary goal of class-incremental learning is to strike a balance between stability and plasticity, where models should be both stable enough to retain knowledge learned from previously seen classes, and plastic enough to learn concepts from new classes.
    While previous works demonstrate strong performance on class-incremental benchmarks, it is not clear whether their success comes from the models being stable, plastic, or a mixture of both.
    This paper aims to shed light on how effectively recent class-incremental learning algorithms address the stability-plasticity trade-off.
    We establish analytical tools that measure the stability and plasticity of feature representations, and employ such tools to investigate models trained with various algorithms on large-scale class-incremental benchmarks.
    Surprisingly, we find that the majority of class-incremental learning algorithms heavily favor stability over plasticity, to the extent that the feature extractor of a model trained on the initial set of classes is no less effective than that of the final incremental model.
    Our observations not only inspire two simple algorithms that highlight the importance of feature representation analysis, but also suggest that class-incremental learning approaches, in general, should strive for better feature representation learning.
\end{abstract}

\section{Introduction}
\input{sections/intro}

\section{Preliminaries}
\input{sections/preliminaries}

%\section{Re-evaluating continually learned feature representations}
\section{Re-evaluating Feature Representations}
\label{sec:plasticity}
\input{sections/investigating_plasticity}

\section{Are Incrementally Learned Features Static?}
%\section{Properties of Incrementally Learned Features}
\label{sec:stability}
\input{sections/stability}

%\section{Improving class-incremental learning based on observations}
\section{Improving Class-Incremental Learning}
\label{sec:exploit}
\input{sections/exploit}

\section{Discussions}
\label{sec:discussion}
\input{sections/discussion}

\section{Conclusion}
\input{sections/conclusion}

%\clearpage

{\small
\bibliographystyle{ieee_fullname}
\bibliography{egbib}
}

%%%%%%%%%%%%%%%%%%%%%%%%%%%%%%%%%%%%%%%%%%%%%%%%%%%%%%%%%%%%

\onecolumn
\pagebreak
\appendix
\maketitle
\renewcommand{\theequation}{\alph{equation}}
\renewcommand{\thetable}{\Alph{table}}
\renewcommand{\thefigure}{\Alph{figure}}

\setcounter{equation}{0}
\setcounter{table}{0}
\setcounter{figure}{0}
%\section{Appendix}
\input{sections/supp_body}

\end{document}

%% file: sections/intro.tex
% !TEX root = ./../main.tex

Despite the unprecedented success of deep learning~\cite{vaswani2017attention, clip,redmon2016you, noh2015learning}, most deep neural networks have static use cases.
However, real-world problems often require adaptivity to incoming data~\cite{nam2016learning}, changes in training environments, and domain shifts~\cite{Ben2010, Ganin2015, dta_iccv2019}.
% Thus, researchers have been actively working on model adaptation techniques and various continual learning algorithms approaches have been proposed so far.
Thus, researchers have been actively working on model adaptation techniques, and have proposed various continual learning approaches so far.

A na\"ive approach for continual learning is to simply fine-tune a model.
However, such a solution is rather ineffective due to a phenomenon known as \textit{catastrophic forgetting}~\cite{kirkpatrick2017overcoming}, which arises as a result of high \textit{plasticity} of neural networks, \ie parameters important for the old tasks are updated to better fit the new data.
On the flip side, enforcing model \textit{stability} introduces its own set of limitations, mainly the lack of adaptivity to new data.
Thus, we encounter the \textit{stability-plasticity dilemma}: how can we balance stability and plasticity such that the model is able to learn new concepts while retaining old ones?
Finding an optimal balance between these two opposing forces is a core challenge of continual learning research, and has been the main focus of many previous works~\cite{rebuffi2017icarl,liu2021adaptive,douillard2020podnet,AFC,yan2021dynamically,park2019continual,park2021class}.

We conduct an in-depth study of recent works in continual learning, specifically concentrating on class-incremental learning (CIL)---a subfield of continual learning---where new sets of classes arrive in an online fashion.
We are motivated by the lack of systematic analyses in the field of CIL, which hampers our understanding of how effectively the existing algorithms balance stability and plasticity.
Moreover, works that do perform analyses usually focus on the classifier, \textit{e.g.,} classifier bias~\cite{SSIL,hou2019learning}, rather than the intermediate feature representations.
%However, investigating the stability and plasticity in the feature level is just as important, if not more, because the capability to learn robust representations by making full use of the model's capacity is critical to maximizing the potential of CIL algorithms.
However, investigating the stability and plasticity in the feature level is just as important, if not more, because utilizing the model's full capacity to learn robust representations is critical to maximizing the potential of CIL algorithms.

To measure plasticity, we retrain the classification layer of CIL models at various incremental stages and study how effectively their feature extractors have learnt new concepts.
We then investigate stability by measuring feature similarity with Centered Kernel Alignment (CKA)~\cite{cka_kornblith, cortes_cka} and by visualizing the feature distribution shift using t-SNE~\cite{tsne}.
Suprisingly, and possibly concerningly, our analyses show that the majority of CIL models accumulate little new knowledge in their feature representations across incremental stages.
In fact, most of the analyzed CIL algorithms seem to alleviate catastrophic forgetting by heavily overlooking model plasticity in favor of high stability.
Finally, we introduce two simple algorithms based on our observations.
The first is an extension of Dynamically Expandable Representations (DER)~\cite{yan2021dynamically}, which demonstrates how our analyses may be used to improve the efficiency and efficacy of CIL algorithms.
%The second is an exploitative method, which, much like GDumb~\cite{prabhu2020gdumb}, raises concerns regarding the current state of CIL research.
The second is an exploitative method, which can be interpreted as an extreme case of DER~\cite{yan2021dynamically} in terms of architectural design; this method shares a similar motivation with GDumb~\cite{prabhu2020gdumb} in the sense that it raises significant concerns regarding the current state of CIL research.
%\dw{
%We establish tools to analyze both the stability and plasticity of existing CIL models.
%As a measure of stability, we propose to adopt Centered Kernel Alignment (CKA)~\citep{cka_kornblith, cortes_cka}, which allows us to observe the similarity of feature representations at each corresponding layer of two models.
%To measure plasticity, we retrain the classifier at various incremental stages and investigate how effectively the feature representations learn new concepts.
%Suprisingly, and possibly concerningly, our analyses show that the majority of CIL models heavily overlook model plasticity in favor of stability.
%To elaborate, the feature representations of most CIL models do not improve as more data is encountered.
%We leverage our observations to develop a very simple (and exploitative) approach, which performs unexpectedly well on the ImageNet class-incremental learning benchmark while requiring no exemplar data and minimal training time.
%Our study suggests that researchers must shift their focus towards actively learning stronger feature representations in order to further advance the field of continual learning research.
%}
%
We summarize our contributions below:
\begin{itemize}
    \item We design and conduct analytical experiments to better understand the balance between stability and plasticity in the feature representations of modern continually learned models. \vspace{-2mm}
    \item We discover that the feature representations of most CIL models are only subject to trivial updates as the model is trained on incremental data. This is a direct result of overweighing the importance of stability over plasticity, and thus, implies a failure of their balance. \vspace{-2mm}
    \item We present two simple but effective CIL algorithms inspired by the results of our analyses. One is an exploit that highlights a major flaw in the current state of CIL research while the other is a variation of an existing method designed for improving performance.
    %\item We present two simple CIL algorithms inspired by the results of our analyses. One is an exploit that highlights a major flaw in the current state of CIL research, while the other improves the efficiency and accuracy of an existing algorithm.
%    \item We present an extremely simple CIL algorithm that exploits the static nature of feature extractors in most models. This exploit is unreasonably effective, which highlights a major flaw in the current state of CIL research.
\end{itemize}

%% file: sections/preliminaries.tex
% !TEX root = ./../main.tex

To set the stage for our paper, we first describe the task setting, notations, and related works.

\subsection{Task setting and notations}

In continual learning, a neural network model is trained on data that arrives incrementally.
Formally, after first training a model with an initial dataset $\data_0$, additional datasets $\{\data_i\}_{i=1}^N$ arrive in $N$ sequential steps to further update the model.
We collectively denote all incremental datasets as $\{\data_i\}_{i=0}^N$ for an $N$-step setting.
In class-incremental learning (CIL), $\data_i$ consists of examples in a set of classes, $\classes_i$, and collectively, $\{\classes_i\}_{i=0}^N$, where all classes are unique such that $|\cup_{i=0}^N \classes_i| = \sum_{i=0}^N |\classes_i|$.
For convenience, we refer to the entire dataset as $\data$, and all classes as $\classes$.
Furthermore, we note that $\data$ may refer to either the training dataset or the validation dataset, depending on the context.

We denote the model trained on $\data_0$ as $\model_0$, and, by extension, define the set of all models trained incrementally on $\{\data_i\}_{i=1}^N$ as $\{\model_i\}_{i=0}^N$.
Note that $\model_i$ ($i > 0$) is first initialized with the parameters of $\model_{i-1}$, and trained on $\data_i$, and optionally with a small exemplar set sampled from $\{\data_i\}_{i=0}^{i-1}$.
We assume the model architectures are based on a convolutional neural network, which, at any given stage $i$, is composed of a feature extractor $\fe_i$, and classifier $\classifier_i$ as
\begin{equation}
    \model_i = \classifier_i \circ \fe_i.
\end{equation}
%
%The classifier typically refers to a single linear layer although it is implemented as a cosine classifier in some algorithms\footnote{\bhr{The details of the cosine classifier are described in the Appendix~\ref{sub:classifier}.}}.
The classifier typically refers to a single linear layer, which may be replaced by a cosine classifier in some algorithms\footnote{The details of the cosine classifier are described in Section~\arxiv{4} of the Appendix.}.

\vspace{-2mm}
\paragraph{Experimental setting}
%\dw{
%Ultimately, the goal of continual learning is to train an incremental model, $\model_N$, such that $\model_N$ achieves strong performance on $\data$.
%However, unlike previous works that report experimental results from $\{\model_i\}_{i=0}^N$, we focus our analyses at the feature extractor, $\{\fe_i\}_{i=0}^N$.
%Thus, in order to eliminate the effects of $\classifier$, we either omit it entirely from the analysis or retrain it on $\data$ while leaving $\fe$ frozen.
%While retraining the classifier on $\data$ is a breach of CIL protocol, we emphasize that our goal is to investigate CIL models from the perspective of feature representations.
%}

To better analyze feature representations in CIL settings, we conduct all experiments on a large-scale dataset, ImageNet-1K~\cite{ILSVRC15}, with a ResNet-18~\cite{he2016deep} architecture, which is a widely-adopted architecture for ImageNet-1K experiments.
There exist three common settings for class-incremental learning on ImageNet-1K:
\vspace{-3pt}
\begin{enumerate}
    \item \textbf{B500-5step}: $|\classes_0| = 500$, $N = 5$, and $|\classes_{i>0}| = 100$.\vspace{-5pt}
    \item \textbf{B500-10step}: $|\classes_0| = 500$, $N = 10$, and $|\classes_{i > 0}| = 50$.\vspace{-5pt}
    \item \textbf{B0-10step}: $|\classes_0| = 100$, $N=9$, and $|\classes_{i > 0}| = 100$.\vspace{-3pt}
\end{enumerate}
%B500-5step setting, where $|\classes_0| = 500$, $N = 5$, and $|\classes_i| = 100, \forall i > 0$, and 2) B500-10step setting, where $|\classes_0| = 500$, $N = 10$, and $|\classes_{i > 0}| = 50, \forall i > 0$.
The first and second settings are both ``pre-trained'' settings, where the model is initially trained on 500 classes, then incrementally updated with fewer classes in later stages.
This setting is quite challenging in terms of stability since the model must retain the pre-trained knowledge when learning from new data.
Moreover, it presents an even greater challenge for plasticity since the model can already extract reasonable feature representations, and thus, must \textit{actively} learn new representations.
We elaborate on this in Section~\ref{sec:discussion}.
Due to space constraints, we present analyses with the B500-5step setting in our main paper, and the latter two in the Appendix.
For all settings, 20 exemplars of each previously seen class is stored in the memory for subsequent stages.

%%%%%%%%%%%%%%%%%
\begin{figure*}[t]
    \centering
    \includegraphics[width=0.9\textwidth]{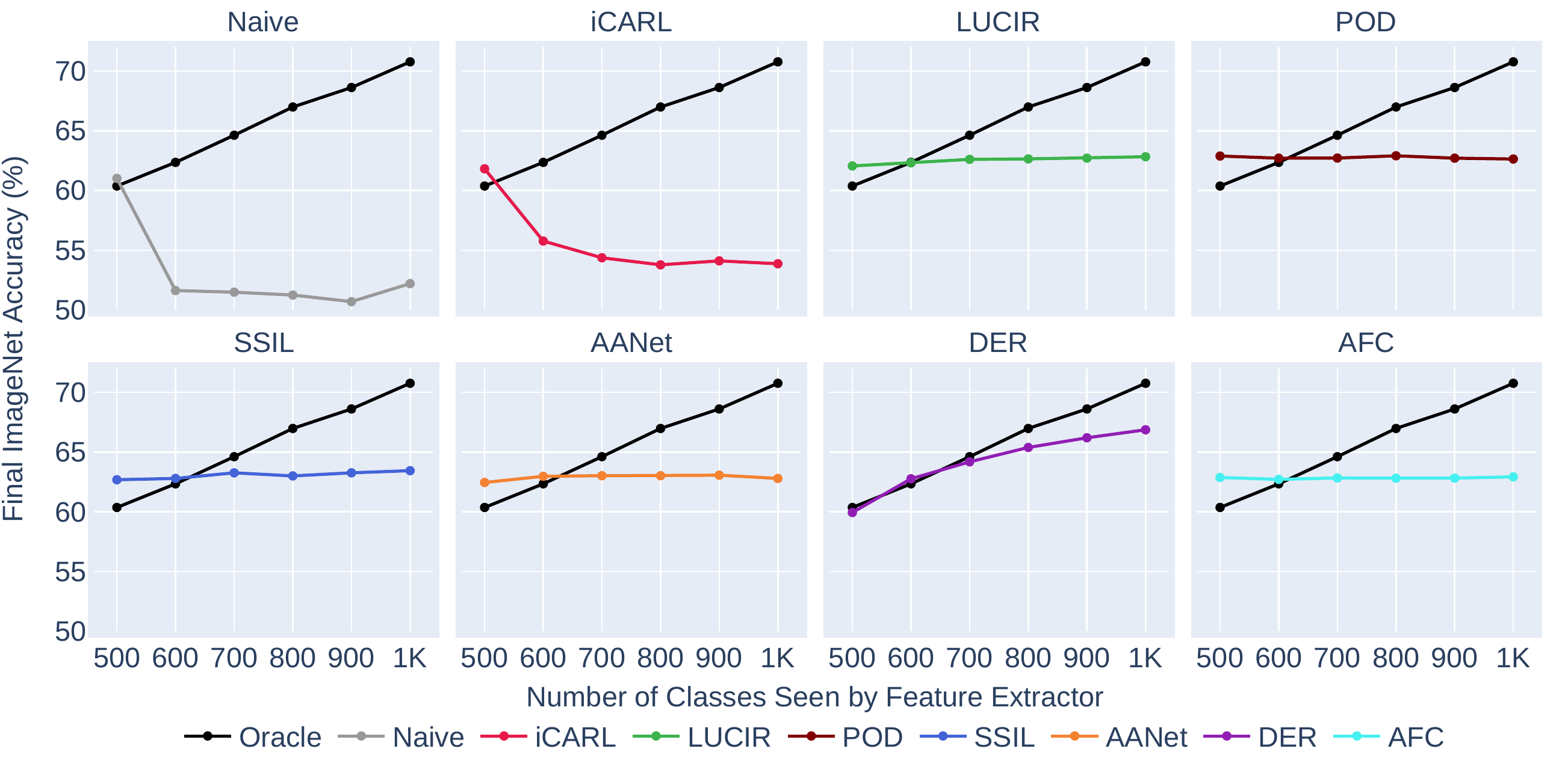}
    \vspace{-3mm}
    \caption{Accuracy on the ImageNet-1K validation set after fine-tuning the classification layer of each incremental model (B500-5step setting) with the full ImageNet-1K training data. The black line indicates an Oracle model trained on \{500, 600, ..., 1000\} classes, and serves as a point of reference for the performance on non-incremental settings.}
    %\vspace{-2mm}
    \label{fig:5step_total_accs}
\end{figure*}
%%%%%%%%%%%%%%%%%

\subsection{Overview of compared methods}
We briefly summarize the main idea of the existing CIL algorithms analyzed in this work.

\vspace{-4mm}
\paragraph{Naive}
%\noindent\textbf{Naive}~~
The naive method adopts simple fine-tuning, where $\model_{i+1}$ is initialized by a fully-trained $\model_i$.
Only the cross-entropy loss is used to train the network, and exemplars are sampled randomly from each previously observed class.

\vspace{-4mm}
\paragraph{iCARL~\cite{rebuffi2017icarl}}
%\noindent\textbf{iCARL~\cite{rebuffi2017icarl}}~~
Incremental Classifier and Representation Learning (iCARL) employs a simple distillation loss alongside the cross-entropy loss. iCARL also proposes herding for exemplar selection, and discovers that nearest-mean-of-exemplars classification can be beneficial for CIL.

\vspace{-4mm}
\paragraph{LUCIR~\cite{hou2019learning}}
%\noindent\textbf{LUCIR~\cite{hou2019learning}}~~
Learning a Unified Classifier Incrementally via Rebalancing (LUCIR) proposes to use the cosine classifier for feature rebalancing purposes, and alleviates the adverse effects of classifier imbalance by using the cosine between features of the student and teacher models.

\vspace{-4mm}
\paragraph{SSIL~\cite{SSIL}}
%\noindent\textbf{SSIL~\cite{SSIL}}~~
Separated Softmax for Incremental Learning (SSIL) identifies that score bias may be caused by data imbalance, and trains the model with a separated softmax output layer alongside task-wise knowledge distillation.

\vspace{-4mm}
\paragraph{AANet~\cite{liu2021adaptive}}
%\noindent\textbf{AANet~\cite{liu2021adaptive}}~~
Adaptive Aggregation Networks (AANet) employs a two branch residual block, where one corresponds to a stable (fixed) block, while the other corresponds to a plastic block.
Existing algorithms can be applied to the AANet architecture.
We focus on AANet + LUCIR, and for simplicity, we denote AANet + LUCIR as AANet.

\vspace{-4mm}
\paragraph{POD~\cite{douillard2020podnet}}
%\noindent\textbf{POD~\cite{douillard2020podnet}}~~
Pooled Outputs Distillation (POD) employs various types of pooling dimensions for knowledge distillation to enforce constraints on feature representations between old and new models.

\vspace{-4mm}
\paragraph{AFC~\cite{AFC}}
%\noindent\textbf{AFC~\cite{AFC}}
Adaptive Feature Consolidation (AFC) first estimates the importance of each channel in the feature map based on the expected increase in loss, and then adaptively restricts the updates to the important channels while leaving non-important channels relatively unconstrained.

\vspace{-4mm}
\paragraph{DER~\cite{yan2021dynamically}}
%\noindent\textbf{DER~\cite{yan2021dynamically}}~~
Dynamical Expandable Representations (DER) adds a new feature extractor at each incremental stage, and leaves feature extractors trained on older data fixed while the new feature extractor is updated.
For any stage $i$, the outputs of all $i+1$ feature extractors are concatenated before passing through the classification layer.

%% file: sections/investigating_plasticity.tex
% !TEX root = ./../main.tex

The lack of model stability and/or plasticity often leads to weak performance, where a suboptimal feature extractor is unable to extract meaningful information.
Likewise, a suboptimal classifier (\textit{e.g.,} due to classifier bias) further exacerbates this issue.
While previous works have extensively studied classifier bias~\cite{SSIL, hou2019learning, zhao2020maintaining}, the effects of unbalanced stability and plasticity in the feature extractor has been relatively less explored, and will be the focus of this section.

%Thus, in order to eliminate the effects of $\classifier$, we either omit it entirely from the analysis or retrain it on $\data$ while leaving $\fe$ frozen.
%While retraining the classifier on $\data$ is a breach of CIL protocol, we emphasize that our goal is to investigate CIL models from the perspective of feature representations.

\subsection{Finetuning the classifier on full data}

%In this section, we take a closer look at model plasticity.
%More specifically, we examine how the feature extractors transform over incremental steps.
We begin by examining how the performance of feature extractors transforms over incremental steps.
%To eliminate any negative effects of an incrementally trained classifier, we freeze the feature extractors $\{\fe_0, ..., \fe_{N}\}$ and train a new classifier for each of the feature extractors on the full ImageNet training data, $\data$.
To eliminate any negative effects of an incrementally trained classifier, we freeze the feature extractors $\{\fe_0, ..., \fe_{N}\}$ and train a new classifier for each of the feature extractors on the full ImageNet-1K training data, $\data$.
In essence, we assume that the classifier is optimal (well-fitted to any given feature extractor), and evaluate the strength of a feature extractor by using the accuracy on the ImageNet-1K validation set as a proxy measure.
Although retraining the classifier on $\data$ is a breach of CIL protocol, we emphasize that the goal of this experiment is purely to analyze CIL models from the perspective of feature representations.

From here on out, $\model'_j \equiv \classifier' \circ \fe_j$ denotes the combination of the feature extractor from stage $j$ and the retrained classifier, $\classifier'$.
In other words, $\fe_j$ is trained incrementally on $\{\data_i\}_{i=0}^{j}$, while $\classifier'$ replaces the original classifier $\classifier_j$ and is retrained on the entire dataset $\data$ with a frozen $\fe_j$.
Note that, while $\classifier_j$ outputs logits for $\sum_{i=0}^j |\classes_i|$ classes, $\classifier'$ outputs $|\classes|$-dimensional logits.
We then define an accuracy metric, \acc{\model'_j}{\data_i}, as the accuracy of model $\model'_j$ on the validation dataset at the $i^\text{th}$ stage, $\data_i$.
Without loss of generality, \acc{\model'_j}{\data} denotes the accuracy of model $\model'_j$ on the full ImageNet-1K validation dataset, $\data$.
%We also define a data efficiency metric, DE, as:
%\begin{equation}
%    \text{DE} = \frac{\text{Acc}(\model^{\text{algorithm}}_N, \classes) - \text{Acc}(\model^{\text{algorithm}}_0, \classes)}{\text{Acc}(\model^{\text{oracle}}_N, \classes) - \text{Acc}(\model^{\text{oracle}}_0, \classes)},
%\end{equation}
%where $\model_j^{\text{oracle}}$ denotes an oracle model, with a feature extractor trained on $\{\data_i\}_{i=0}^{j}$ and a classifier retrained on $\data$.
%DE measures the data efficiency of incremental feature extractors with respect to an oracle feature extractor, \textit{i.e.,} how much the feature extractor improves as training classes are added.

%%%%%%%%%%%%%
\begin{figure*}[t]
    \centering
    \includegraphics[width=0.9\textwidth]{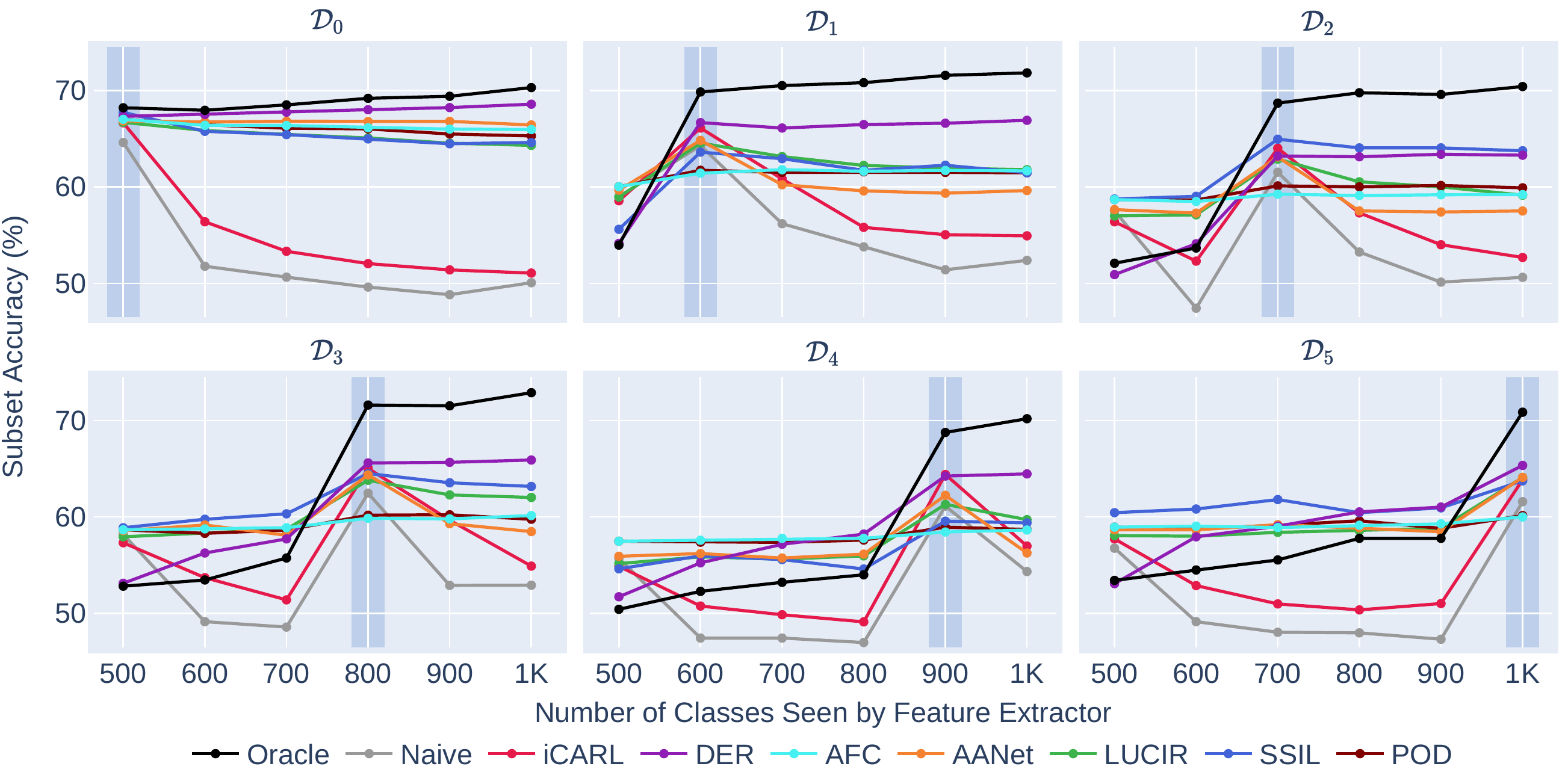}
    \vspace{-2mm}
    \caption{B500-5step subset accuracies. For the sake of visibility, we leave out SSIL and POD from these plots.
    We highlight the region for model $\model'_j$ in plot $\classes_i$, where $i = j$.}
    \label{fig:5step_subset_accs}
    \vspace{-2mm}
\end{figure*}
%%%%%%%%%%%%%

%\vspace{-2mm}
%\paragraph{ImageNet-1K accuracies}
\subsection{ImageNet-1K accuracies}

Figure~\ref{fig:5step_total_accs} illustrates the full validation accuracy on ImageNet-1K, \acc{\model'_j}{\data}, for all the compared algorithms.
Each subplot visualizes the accuracy progression for the specified CIL algorithm as well as the Oracle model, $\model_{j}^*$, whose feature extractor is trained on $\cup_{i=0}^j \data_i$ all at once, before the classifier is retrained on $\data$.
In essence, the feature extractor of $\model_j^*$ represents how an ideal incremental model would perform if important features from previous tasks are not forgotten and new concepts are well learned.

Figure~\ref{fig:5step_total_accs} generally exhibit three distinct trends.
First, we observe the plots for the Naive and iCARL models, and notice that the accuracy significantly declines after the initial stage, \textit{i.e.,} \acc{\model'_0}{\data} $>$ \acc{\model'_{j}}{\data}, $\forall j > 0$.
These results imply that both models are subject to severe catastrophic forgetting, and their ability to extract useful features deteriorates significantly from the first incremental stage.
%\dw{Along with the low CKA values for the naive and iCARL models in Figure~\ref{fig:cka_5step}, these results suggest that both the naive and iCARL models are subject to severe catastrophic forgetting.}
Next, we investigate the subplots for LUCIR, POD, SSIL, AANet, and AFC.
Surprisingly, for these five distinct CIL algorithms, the accuracy remains almost unchanged across all incremental stages, despite a few minor variations.
Unlike the Oracle model, which shows an almost linear increase in accuracy at each incremental step, these five CIL algorithms maintain the same accuracy, \textit{i.e.,} \acc{\model'_0}{\data} $\approx$ \acc{\model'_j}{\data}, $\forall j > 0$.
Thus, for these 5 algorithms, the feature extractor of $\model'_5$ is not particularly stronger than the feature extractor of $\model'_j$, $\forall j < 5$.
Lastly, DER exhibits increasing accuracy with each incremental stage, indicating that the feature extractor does indeed learn new features.

%%%%%%%%%%%%%%%%%%
\begin{figure*}[t]
    \centering
    \includegraphics[width=0.95\textwidth]{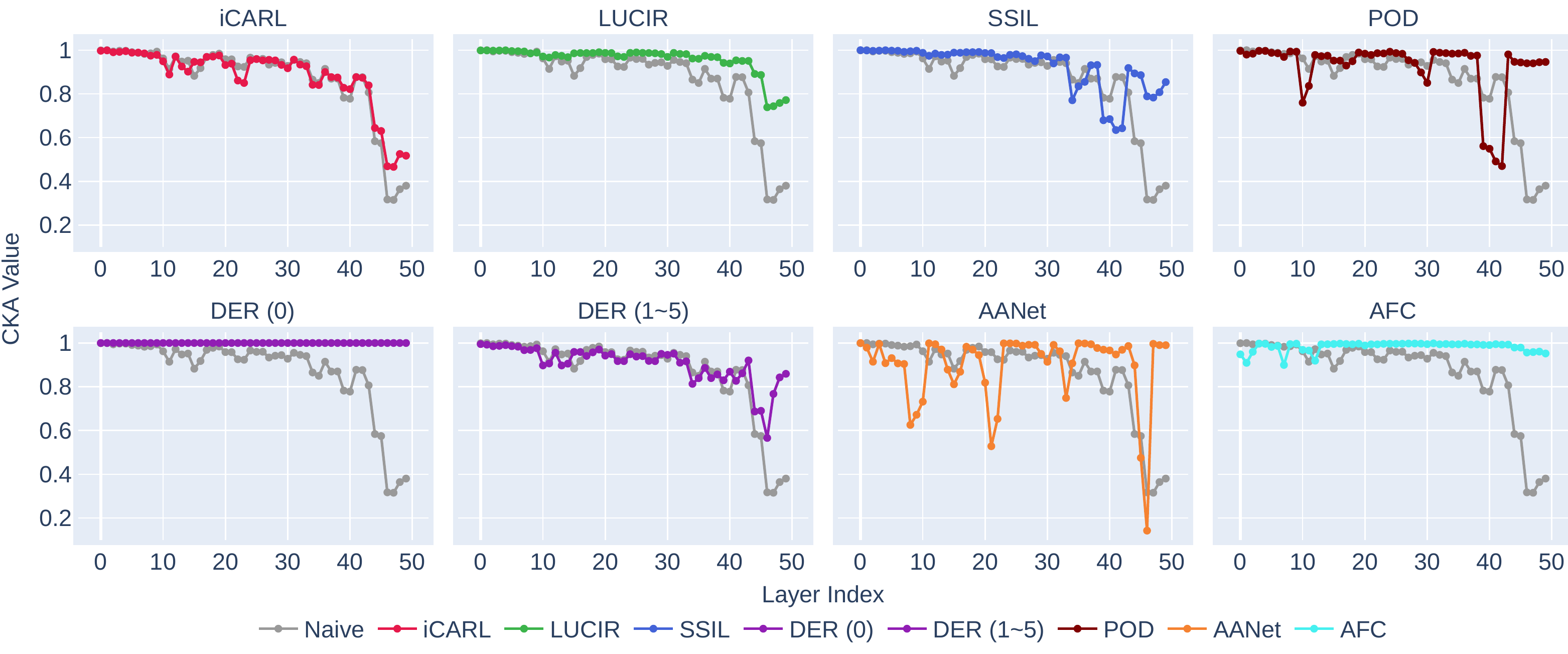}
    \vspace{-2mm}
    \caption{Same-layer CKA values between $\fe_0$ and $\fe_5$ for incremental models trained with each CIL algorithm on the ImageNet B500 5-step setting. The x-axis spans the layer index of ResNet-18, while the y-axis represents CKA. CKA is evaluated using the $\data_0$ validation set. Each plot is accompanied by the CKA for the naive model, which acts as a point of reference.}
    \label{fig:cka_5step}
    \vspace{-2mm}
\end{figure*}
%%%%%%%%%%%%%%%%%%

%%%%%%%%%%%%%%%
%\vspace{-2mm}
%\paragraph{ImageNet-1K subset accuracies}
\subsection{ImageNet-1K subset accuracies}

Diving deeper, we investigate how the accuracy of each subset, $\data_i$, changes at each incremental stage.
Figure~\ref{fig:5step_subset_accs} illustrates the change in \acc{\model'_j}{\data_i}, $\forall j$ for each $i$.
For the sake of better visibility, we omit POD and SSIL in this figure and present the full plots in the Appendix.

First, we focus on the gray curve, representing the Naive model.
We expected the Naive model to suffer from catastrophic forgetting, and overfit to the most recently seen set of classes.
Indeed, the results corroborate our intuition; in the first plot on $\data_0$, \acc{\model'_j}{\data_0}, $\forall j > 0$, performs significantly worse than \acc{\model'_0}{\data_0}.
%Furthermore, in the subplots for $\data_1 \sim \data_5$, we notice that \acc{\model'_j}{\data_i} peaks when $j = i$.
Furthermore, we notice that \acc{\model'_j}{\data_i} peaks when $j = i$.
Then, \acc{\model'_j}{\data_i} drops off again when $j > i$.
These observations all lead to the same conclusion that the naive model suffers from catastrophic forgetting due to its high plasticity.
A similar pattern is observed from the models trained by iCARL.

Next, we shift our focus to the black curve corresponding to the Oracle model.
We notice that for $\data_0$, \acc{\model'_j}{\data_0} does not drop, but rather increases as more classes are added, which suggests that knowledge from $\classes_i$, $\forall i > 0$, can in fact be beneficial for performance on $\classes_0$.
Moreover, \acc{\model'_j}{\data_i} significantly increases when $j = i$.
Altogether, the trends exhibited by the Oracle model represent what an ideal CIL model would demonstrate.
%Interestingly, we notice that \acc{\model'_j}{\data_i} does not necessarily increase when $j = i$; in some cases (such as $\data_4$), \acc{\model'_j}{\data_i} $>$ \acc{\model'_k}{\data_i}, for $k > j$.
%While DER follows a similar trend to the oracle --- where the \acc{\model'_j}{\data_0} increases with $j$ --- we also note that \acc{\model'_j}{\data_i} always significantly increases when $j = i$.

Finally, we look at the cyan, orange, and green curves, respectively representing AFC, AANet, and LUCIR.
For AFC, \acc{\model'_j}{\data_i} is mostly unchanged $\forall i, j$, suggesting that their features are mostly static across all incremental stages.
While this implies that no forgetting occurs, it appears to come at the cost of learning little to no new concepts.
Meanwhile, both AANet and LUCIR are mostly stable, but also slightly more plastic than AFC and POD; they quickly forget newly learned concepts since \acc{\model'_j}{\data_i} peaks when $j = i$ but drops back down when $j > i$.

%% file: sections/stability.tex
% !TEX root = ./../main.tex
%\dw{
%We begin by investigating the extent to which incremental models, trained on modern CIL algorithms, mitigate catastrophic forgetting on the feature level.
%One simple method of measuring forgetting is to calculate the performance degradation of each previous set of classes as new classes arrive.
%However, the results of such analysis can be heavily affected by the classifier, which may have varying degrees of optimality.
%We take a different approach and analyze the intermediate feature representations of an incremental model.
%More specifically, in this section, we first establish a tool to measure how the feature representation at any given layer changes across incremental stages, and employ it to analyze models trained on recent CIL algorithms.
%}
Our analysis in Section~\ref{sec:plasticity} suggests that a majority of the compared CIL algorithms appear to have high feature stability at the cost of low plasticity.
This raises a question: do feature representations remain static across incremental models?
In hopes to shed light on this issue, we measure the similarity of intermediate activations and visualize the feature distribution shifts between incremental models.

%%%%%%%%%%%%%%%%%%
\subsection{Centered Kernel Alignment (CKA)}
To analyze intermediate representations of a neural network, we employ Centered Kernel Alignment (CKA)~\cite{cka_kornblith, cortes_cka}, which enables us to quantify the similarity between pairs of neural network representations.
CKA has been used to study the effects of increasing depth and width in a network~\cite{dodeepandwide},
and to understand how the representations of Vision Transformers~\cite{vit} differ from those of convolutional neural networks~\cite{CKADoVisionTransformers}.

Let us consider two arbitrary layers of a neural network with $z_1$ and $z_2$ output features.
Given the same set of $b$ inputs, we denote the activation matrices as $\mathbf{X} \in \mathbb{R}^{b \times z_1}$ and $\mathbf{Y} \in \mathbb{R}^{b \times z_2}$.
The $b \times b$ Gram matrices $\mathbf{K} = \mathbf{X}\mathbf{X}^T$ and $\mathbf{L} = \mathbf{Y}\mathbf{Y}^T$ are first centered to obtain $\mathbf{K'}$ and $\mathbf{L'}$, which are then used to compute the Hilbert-Schmidt Independence Criterion (HSIC)~\cite{HSIC} as
\begin{equation}
    \text{HSIC}(\mathbf{K}, \mathbf{L}) = \frac{\text{vec}({\mathbf{K'}}) \cdot \text{vec}(\mathbf{L'})}{(b - 1)^2},
\end{equation}
where $\text{vec}(\cdot)$ denotes the vectorization operation.
Finally, CKA normalizes HSIC as follows:
\begin{align}
    \label{eq:CKA}
    \text{CKA}(\mathbf{X}, & \mathbf{Y}) = \\
    &\frac{\text{HSIC}(\mathbf{XX^\top}, \mathbf{YY^\top})}{\sqrt{\text{HSIC}(\mathbf{XX^\top}, \mathbf{XX^\top}) \text{HSIC}(\mathbf{YY^\top}, \mathbf{YY^\top})}}. \nonumber
\end{align}
As shown above, CKA is a normalized measure of how similar the $b \times b$ Gram matrices $\mathbf{K}$ and $\mathbf{L}$ are.
Given that the Gram matrices themselves reflect the feature relationships among pairs of samples, CKA can be interpreted as a similarity of relationships among the features in $\mathbf{X}$ and $\mathbf{Y}$.

In terms of comparing feature representations, we highlight three properties that make CKA stand out.
First, CKA is invariant to permutations in the columns of $\mathbf{X}$ and $\mathbf{Y}$, \textit{i.e.,} $\text{CKA}(\mathbf{X}, \mathbf{Y}) = \text{CKA}(\mathbf{XP}, \mathbf{Y})$, where $\mathbf{P} \in \{0, 1\}^{z_1 \times z_1}$ is an arbitrary permutation matrix. Second, it is invariant to isotropic scaling of $\mathbf{X}$ and $\mathbf{Y}$.
Finally, it can be used to compare activations of layers with different output feature dimensions, \textit{e.g.,} different layers of the same network, or even layers across different architectures.
Such advantages make CKA suitable for analyzing the feature representations of class-incremental models.
%which we outline in the upcoming section.
%%%%%%%%%%%%%%%%%%

%%%%%%%%%%%%%%%%%%
%\subsection{Measuring the similarity of incremental feature representations with CKA}
\subsection{Measuring the similarity of representations}
\label{sec:cka}
%With the correct experimental setup, CKA can be adopted to measure the stability of features across incremental models in the class-incremental setting.
%Intuitively, a model that is stable across incremental  will extract similar intermediate feature representations across all s
Given a pair of incremental feature extractors $\fe_0$ and $\fe_{N}$ trained by a selected class-incremental algorithm ($N = 5$ for the B500-5step setting and $N=10$ for the B500-10step setting), we extract a set of features $\{\mathbf{X}_l\}_{l=0}^{L}$ from $\fe_0$, where $\mathbf{X}_l$ denotes feature output of layer $l$.
After extracting a corresponding set of features $\{\mathbf{Y}_l\}_{l=0}^{L}$ from $\fe_{N}$, we compute CKA\footnote[1]{More precisely, we use the mini-batch version of CKA, which has been shown to converge to the same value of full-batch CKA, described in Eq. (\ref{eq:CKA})~\cite{dodeepandwide}. We provide details for mini-batch CKA in the Appendix.} between $\mathbf{X}_l$ and $\mathbf{Y}_l$, $\forall l \in \{0,\dots, L\}$, using the ImageNet-1K $\data_0$ validation subset.
We extract activations from all convolution, batch normalization~\cite{ioffe2015batch}, and residual block layers of ResNet-18, which results in two sets of features, each with a cardinality of 50 ($L = 49$).
%We hypothesize, from simple intuition, that a high CKA between $\mathbf{X}_l$ and $\mathbf{Y}_l$, \textit{i.e.,} $\text{CKA}(\mathbf{X}_l, \mathbf{Y}_l) \approx 1$, can serve as a strong indicator for \textit{low forgetting} by virtue of $\mathbf{X}_l$ and $\mathbf{Y}_l$ being highly similar.
A high CKA between $\mathbf{X}_l$ and $\mathbf{Y}_l$, \textit{i.e.,} $\text{CKA}(\mathbf{X}_l, \mathbf{Y}_l) \approx 1$, indicates that the two feature representations are highly similar, and thus, remained static across incremental stages.
By extension, high CKA may also serve as a strong indicator for \textit{low forgetting} due to high similarity of $\mathbf{X}_l$ and $\mathbf{Y}_l$.

Figure~\ref{fig:cka_5step} presents the same-layer CKA between $\fe_0$ and $\fe_{5}$ trained on various CIL algorithms.
Each subplot visualizes the layer-wise CKA for two algorithms: 1) the Naive method, which represents a fully-plastic baseline, and 2) the specified CIL algorithm.
We first inspect the naive model, which displays relatively high CKA in early layers, but significantly deteriorates in latter layers.
This observation is consistent with the notion that early layers learn low-level and widespread features, while higher layers tend to learn more class-specific information~\cite{Zeiler_visualizingNN}.
Then, a cursory examination of all other algorithms suggest that all compared CIL algorithms do indeed enforce feature representations to be similar across incremental models, albeit with varying levels of success.
For example, both POD and AANet retain high feature similarities across most layers, although there are some significant drops in a few select layers (more details on CKA of AANet provided in the Appendix).
Furthermore, we observe that AFC maintains high CKA across all layers, implying that the model trained with AFC has high stability, and thus, undergoes little to no forgetting in each incremental step.
%However, as we will explore in Section~\ref{sec:plasticity}, such high stability comes at the expense of a reduced ability to learn features for newer classes.

Finally, we present two plots for DER.
In the B500-5step setting, DER consists of 6 separate feature extractors, the first of which is identical to $\fe_0$.
Thus, we plot CKA separately for the fixed feature extractor, DER~(0), and all other feature extractors, DER~(1$\sim$5).
As expected, we observe maximum CKA for DER~(0), and much lower CKA for DER~($1 \sim 5$).

\subsection{t-SNE Visualizations}
%\begin{figure*}[t]
%    \centering
%%    \includegraphics[width=\columnwidth]{figures/tsne_last_stage_classes}
%    \includegraphics[width=0.8\textwidth]{figures/tsne_last_stage_classes}
%%    \vspace{-5mm}
%    \caption{t-SNE visualization of 5 classes of $\classes_5$, with 20 samples from each class. Each color represents a distinct class, and features of $\fe_0$ and $\fe_5$ are depicted with a circle and star, respectively.}
%    \label{fig:tsne}
%%    \vspace{-2mm}
%\end{figure*}

\begin{figure*}[t]
    \centering
    \begin{tabular}{cc}
        \multicolumn{2}{c}{\includegraphics[width=0.8\textwidth]{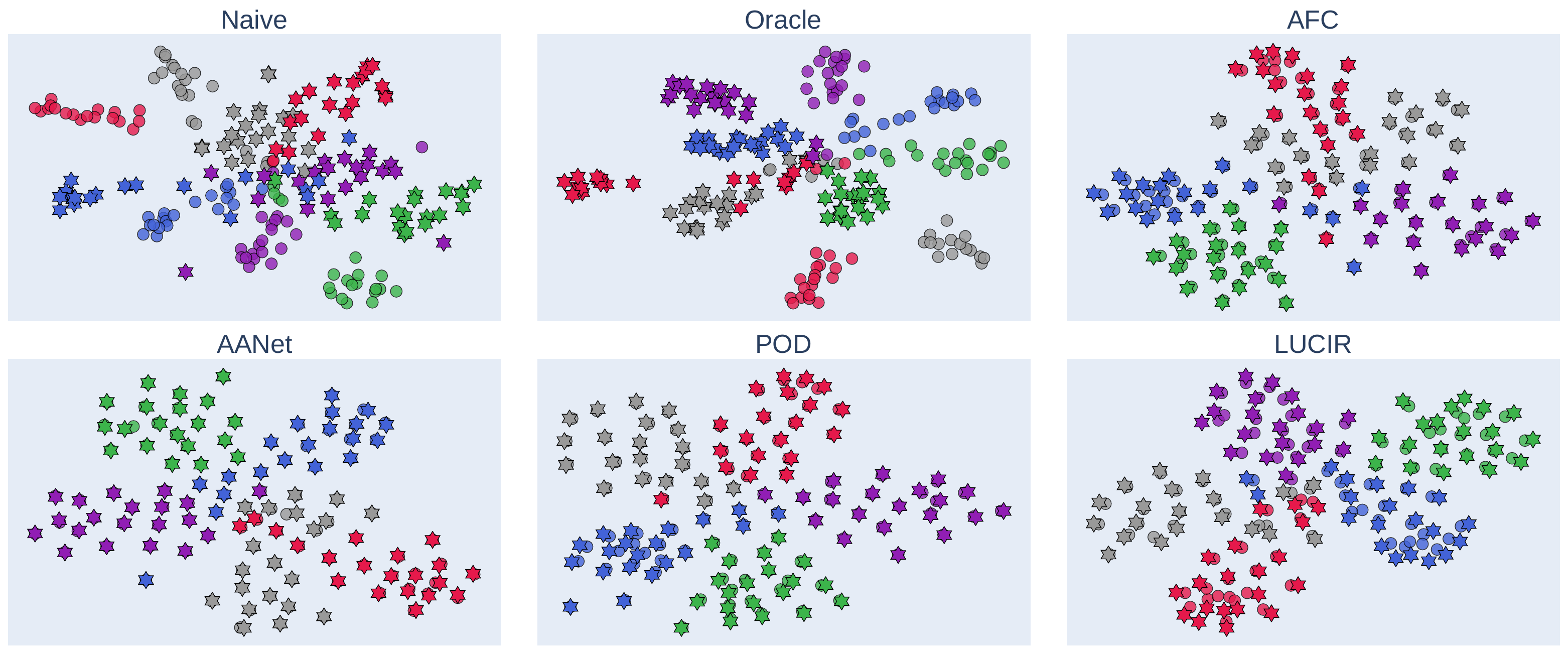}}\vspace{-1mm}\\
        \hspace{6mm}\includegraphics[width=0.35\textwidth]{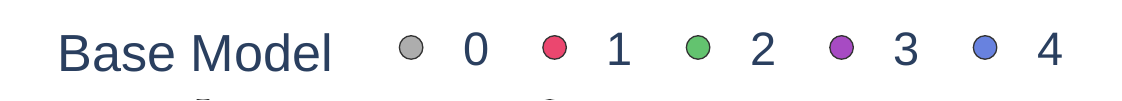} & \includegraphics[width=0.35\textwidth]{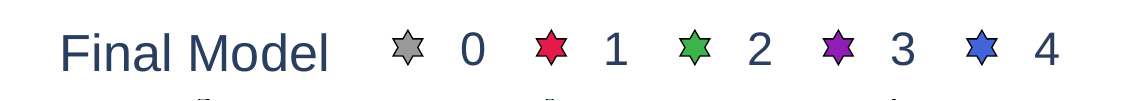}

    \end{tabular}
    \vspace{-3mm}
    \caption{t-SNE visualization of features from $\fe_0$ and $\fe_5$ using 5 classes of $\classes_5$ and 20 samples from each class. All 200 embeddings are visualized in a single plot. Each color represents a distinct class, and features of $\fe_0$ and $\fe_5$ are depicted with a circle and star, respectively.}
    \label{fig:tsne}
\end{figure*}

To further corroborate our observations from Section~\ref{sec:cka}, we visualize the feature shift t-SNE~\cite{tsne}.
We randomly sample 5 classes from $\classes_5$ and 20 images from each selected class.
Then, we compute the feature representations of each image using both $\fe_0$ and $\fe_5$, and visualize all 200 features in a single t-SNE plot.

Figure~\ref{fig:tsne} illustrates the t-SNE plots of the Naive and Oracle feature extractors, as well as those of four CIL algorithms that exhibit high stability.
In the Naive and Oracle plots, we observe that the same-class features have shifted significantly, since the $\fe_0$ and $\fe_5$ features are clustered in different regions.
This is expected since both models exhibit high plasticity.
In the AFC, AANet, POD, and LUCIR plots, however, the feature shift is trivial.
Interestingly, most of the $\fe_0$ and $\fe_5$ features corresponding to the same inputs are overlapped, even though $\fe_5$ has been trained on $\{\data_i\}_{i=1}^{5}$, while $\fe_0$ has not.
Once again, this suggests that these algorithms excel at maintaining important knowledge from previous tasks, but fail to acquire new knowledge from incremental data.

%\input{tables/summary}
%\input{tables/cil_benchmarks}

%% file: sections/exploit.tex
% !TEX root = ./../main.tex

%\subsection{Exploiting static feature extractors \dw{MOVE ELSEWHERE}}
We now outline two methods to improve CIL models, inspired by the analyses presented in Sections~\ref{sec:plasticity} and~\ref{sec:stability}.

\subsection{Partial-DER}
\label{sec:partial_der}
Figures~\ref{fig:5step_total_accs} and~\ref{fig:5step_subset_accs} show that DER not only retains knowledge from old classes, but is also adept at learning concepts from new classes.
Such properties of DER are desirable for CIL algorithms.
DER achieves this by freezing and maintaining $\{\fe_j\}_{j=0}^{i-1}$ at the $i^\text{th}$ incremental stage, and adding a new fully plastic feature extractor (with the same number of parameters as $\fe_{0}$) for the current set of classes.
This allows the feature extractor to fully learn concepts of new classes all the while maintaining knowledge of old classes.
However, a major issue of DER is scalability, since for an $N$-step CIL model, a single forward pass through $\fe_N$ requires $N+1$ times computation compared to $\fe_0$.

Based on our CKA analysis in Figure~\ref{fig:cka_5step}, we propose a modification of DER, called partial-DER (pDER), which not only makes DER much more efficient but also improves the overall performance as well.
In Section~\ref{sec:cka}, we observed that even the Naive model maintains high feature similarity in the lower layers of the network, \textit{i.e.,} the feature representations of lower layers do not change much across $\fe_0$ and $\fe_5$ even when the model is fully plastic.
This suggests that lower layers are inherently stable.
Thus, instead of maintaining $N$ full feature extractors for an $N$-step setting, pDER fixes the lower subset of layers in $\fe_0$, and only applies DER on the upper subset of layers.
More specifically, we consider all layers up to ResNet's Layer 4 as the lower subset, and apply DER only for Layer 4.
We find that this simple modification reduces the GMACs of a forward pass through a DER model by up to 65\%\footnote{$\fe_5$ of DER requires 10.9 GMACs for one forward pass while $\fe_5$ of pDER does 3.9 GMACs. More details are provided in the Appendix.}, while improving \acc{\model'_5}{\data} and \acc{\model_5}{\data} by 1.5\%p and 0.9\%p, respectively.

\subsection{Exploiting static feature extractors}
\label{sec:static_exploit}
%Based on our observations from Sections~\ref{sec:plasticity} and~\ref{sec:stability}, we develop a simple class-incremental exploit that performs surprisingly well on the ImageNet B500-5step setting.
%The purpose of this exploit is not to propose a novel method, nor is it to achieve state-of-the-art performance.
%Rather, our goal is to demonstrate that we can abuse the static nature of feature extractors in CIL models to train a model that requires a fraction of the training time, uses no exemplars nor any special data augmentation, yet is still competitive on the ImageNet B500-5step benchmark.
The second method, which we name ``Exploit'', is based on the observation that the feature extractors in most CIL models remain static over the course of incremental stages.
In such case, we can significantly improve the training efficiency, achieve strong performance, and eliminate the need for previous-class exemplars by simply freezing the base feature extractor, $\fe_0$.
This exploit can be interpreted as an extreme case of pDER, where branching occurs in the classification layer.

We first train $\model_0$ on $\data_0$, and fix the feature extractor, $\fe_0$, for all subsequent incremental steps.
%Fixing the feature extractor allows us to tune the parameters of a classifier independently of classes from previous stages.
The weight matrix of the cosine classifier $\classifier_0$ is denoted as $\mathbf{W}_0 \in \mathbb{R}^{F \times |\classes_0|}$, where $F$ is the output feature dimension of the feature extractor.
For each incremental step $i$, we train a new weight matrix for the classifier, $\mathbf{W}_i \in \mathbb{R}^{F \times |\classes_i|}$, where the cross-entropy loss is only computed for $\classes_i$, \textit{i.e.,} the softmax operation only considers the logits for $|\classes_i|$ classes.
Since $\fe_0$ is fixed for all stages, we do not need to update $\mathbf{W}_j, \forall j < i$.
After training on $\data_N$, we concatenate the set of weight matrices $\{\mathbf{W}_i\}_{i=0}^N$ to produce a single weight matrix, $\mathbf{W} \in \mathbb{R}^{F \times |\classes|}$.
Finally, we compose $\classifier$, parametrized by weight $\mathbf{W}$, with $\fe_0$ to obtain the final model: $\model_{N} = \classifier \circ \fe_0$.

Compared to CIL algorithms that usually train $\model$ for 90 to 120 epochs on each $\data_i$, our exploit only requires around 10 epochs of training to converge for each incremental stage.
Furthermore, we only need to compute gradients for $\classifier$, which reduces the computational burden of training.
These two factors make training extremely fast compared to traditional algorithms that tune the entire model for many epochs.
Despite requiring only a fraction of the computation, our exploit achieves final ImageNet-1K accuracy of 60.5\% and an average incremental accuracy of 67.2\%, which are competitive against the other methods.

%% file: sections/discussion.tex
% !TEX root = ./../main.tex
%\subsection{Quantitative summary}
Based on the plots in Figure~\ref{fig:5step_total_accs}, we define a metric as
\begin{equation}
  \label{eq:data_efficiency}
%    \text{DE} := \frac{\text{Acc}(\model'^{\text{CIL}}_N, \data) - \text{Acc}(\model'^{\text{CIL}}_0, \data)}{\text{Acc}(\model'^{\text{oracle}}_N, \data) - \text{Acc}(\model'^{\text{oracle}}_0, \data)}.
  \Delta\model'_i := \text{Acc}(\model'_i, \data) - \text{Acc}(\model'_0, \data),
\end{equation}
which measures the relative performance of improvement of $\fe_i$ over $\fe_0$, \textit{i.e.,} how much the feature extractor improves as training classes are added.
A high positive $\Delta\model'_i$ indicates that the feature extractor is able to learn new concepts incrementally.
On the other hand, $\Delta\model'_i \simeq 0$ indicates that the feature extractor is stable, but not plastic enough to acquire new knowledge.
Finally, a large negative $\Delta\model'_i$ represents severe catastrophic forgetting in the feature extractor.
%DE value of $1$ represents an incremental model with the same data efficiency as training on all $\data$ at once.
%DE can also be negative, if the feature extractor of $\model_N$ becomes weaker as the model is trained on more data.
%Furthermore, near zero DE may indicate one of two scenarios: 1) the incremental model does is stable and non-plastic, or 2) the amount of forgetting is offset by an equal amount of new concept learning.
Ultimately, CIL models should strive to maximize $\Delta\model'_i$ in order to facilitate better feature representation learning.

\input{tables/summary}

\input{tables/cil_benchmarks}

\vspace{-2mm}
\paragraph{Are we learning continually?}
Table~\ref{tab:summary} quantitatively summarizes all the compared CIL algorithms in terms of \acc{\model'_0}{\data}, \acc{\model'_5}{\data}, and $\Delta\model'_5$.
Looking at LUCIR, SSIL, POD, AANet, and AFC in Table~\ref{tab:summary}, we notice that there is almost no difference between \acc{\model'_0}{\data} and \acc{\model'_5}{\data}, which all lie in the range of $62\%\sim 63\%$.
Furthermore, there exists a significant gap between \acc{\model'_5}{\data} of the aforementioned algorithms and \acc{\model'_5}{\data}$=70.8\%$ of the Oracle model.
This implies that, while these methods do mitigate catastrophic forgetting to varying degrees, they suffer from a lack of plasticity, \textit{i.e.,} are unable to learn continually.
Moreover, these results also imply that if feature representations do not accumulate new knowledge through incremental data, the highest achievable performance is still 7$\sim$8\%p shy of the Oracle model even with an optimal classifier.
Thus, the plasticity of feature representations is an absolutely crucial aspect of modern incremental learning algorithms.

\vspace{-2mm}
\paragraph{Effect of task similarity}
Task similarity has been shown to affect the amount of catastrophic forgetting that neural networks experience~\cite{NTKOverlap2021}.
Despite this, we find that our analyses lead to the same conclusions even when the level of task similarity changes across stages.
%To investigate whether the conclusions made from our analyses are consistent even when the level of task similarity changes, we conduct experiments with lower task similarity between stages.
These experimental results can be found in Section 5 of the Appendix.

\vspace{-2mm}
\paragraph{Pretrained setting}
At this point it is reasonable to question whether the lack of plasticity may be attributed to the experimental setting, \ie B500-5step, where models are pretrained with 500 classes before being updated incrementally.
While this may be true to some extent\footnote{Analysis of select algorithms on B0-10step included in the Apendix.}, we argue that continual learners should strive towards obtaining better feature representations \textit{even if} the pretrained representations are somewhat sufficient.
If not, what good is a continual model compared to a static one as described in Section~\ref{sec:static_exploit}?
In fact, we have shown that the pretrained feature extractor is still lackluster compared to the Oracle while DER has shown that, albeit lacking scalability, it is possible to continually learn stronger representations even with a pretrained feature extractor.
Thus, our choice to conduct analysis on this pretrained setting cannot justify the lack of plasticity exhibited by most CIL algorithms.

%Works with rigorous theory suggest that forgetting can be minimized whilst learning new information (\eg via orthogonal projection~\cite{NTKOverlap2021}), but since they only evaluate on smaller datasets (\eg MNIST), it is unclear whether these methods remain effective for large-scale datasets (\eg ImageNet).

\vspace{-2mm}
\paragraph{Limitations of traditional metrics}
In Table~\ref{tab:cil} we present the widely adopted metrics in CIL: 1) average incremental accuracy of all compared CIL algorithms on the B500-5step setting and 2) the final accuracy on ImageNet, \acc{\model_5}{\data}.
In particular, we note that our Exploit in Table~\ref{tab:cil} outperforms all but one CIL algorithm (DER) on both metrics, despite keeping the feature extractor $\fe_0$ fixed.
%Furthermore, LUCIR and AANet demonstrate weak performance here, as their average incremental and final ImageNet-1K accuracies are roughly 10\%p and 20\%p lower than POD, AFC, DER, SSIL, and Exploit.
However, according to Table~\ref{tab:summary}, all the aforementioned CIL algorithms demonstrate similar \acc{\model'_0}{\data} and \acc{\model'_5}{\data} scores.
This suggests that while the feature representations of all compared methods except Naive, iCARL, and DER have similar levels of discriminativeness, it is not well expressed in terms of average incremental accuracy nor \acc{\model_5}{\data}.
%This raises significant concerns regarding the way we evaluate CIL algorithms; c
Clearly, high average incremental accuracy and final accuracy are not really indicative of how much the model has \textit{learned} continually; yet, these metrics have become the de-facto standard in CIL research.
This should be alarming for both researchers and practitioners;
ambiguous metrics deliver a false illusion of progress, and may lead researchers to develop algorithms that \textit{seem} to outperform other state-of-the-art algorithms, but are completely misaligned with the motivation behind continual learning, \eg our Exploit.
%such as our Exploit.
Thus, we hope that the analyses in our work will facilitate better evaluation of CIL algorithms and inspire researchers to focus more on stronger feature representation learning of incremental models.

\vspace{-2mm}
\paragraph{Connection with theory}
Of the works we study, the only algorithm that exhibits both high plasticity and stability is based on parameter isolation, which makes us wonder whether it is even possible to achieve a strong balance of stability and plasticity without adding new model parameters.
However, Raghavan~\etal~\cite{CLSequentialGame21} prove that there can exist a stable equilibrium point between forgetting and generalization.
In fact, it is possible that the low-plasticity methods do indeed reach a stable equilibrium, except that this equilibrium is less optimal in terms of performance.
Therefore, a key question for future works is: how can we reach a \textit{more optimal} equilibrium point at each incremental stage without adding more parameters?

%% file: tables/summary.tex
\begin{table*}
  \caption{Summary of all compared CIL algorithms. *Methods introduced in Section~\ref{sec:exploit}}
  \label{tab:summary}
  \vspace{-2mm}
  \centering
  \setlength{\tabcolsep}{5pt}
  \renewcommand{\arraystretch}{1.1}
  \scalebox{0.9}{
  \begin{tabular}{lccccccccccc}
    \toprule
%    \multicolumn{2}{c}{Part}                   \\
%    \cmidrule(r){1-2}
    Method & Naive & iCARL & LUCIR & SSIL & POD & AANet & AFC & DER & Exploit* & pDER* & Oracle \\
    \midrule
%    Final CKA & 0.380 & 0.517 & 0.772 & 0.854 & 0.946 & 0.990 & 0.952 & 1.000 & 1.000 & - \\
%    Mean CKA & 0.873 & 0.884 & 0.957 & 0.926 & 0.926 & 0.894 & 0.983 & 1.000 & 1.000 & - \\
    \acc{\model'_0}{\data} & 60.4 & 61.8 & 62.1 & 62.5 & 62.9 & 62.5 & 62.9 & 60.0 & 62.9 & 60.0 & 60.4 \\
    \acc{\model'_5}{\data} & 52.2 & 53.9 & 62.8 & 63.5 & 62.6 & 62.8 & 62.9 & 66.9 & 62.9 & 68.4 & 70.8 \\
    $\Delta\model'_5$ & -8.2 & -7.9 & 0.7 & 1.0 & -0.3 & 0.3 & 0.0 & 6.9 & 0.0 & 8.4 & 10.4 \\
%    DE (Eq.(~\ref{eq:data_efficiency})) & -0.85 & -0.76 & 0.07 & 0.10 & -0.03 & 0.03 & 0.00 & 0.66 & 0.00 & 0.81 & 1.00 \\
%    Lack of: & Stab. & Stab. & Plast. & Plast. & Plast. & Plast. & Plast. & Scal. & Plast. & - \\
    \bottomrule
  \end{tabular}
  }
  \vspace{-1mm}
\end{table*}

%\begin{table}
%  \caption{Summary of all compared CIL algorithms. Stab. = Stability, Plast. = Plasticity, Scal. = Scalability}
%  \label{tab:summary}
%  \centering
%  \setlength{\tabcolsep}{5.5pt}
%  \renewcommand{\arraystretch}{1.1}
%  \scalebox{0.9}{
%  \begin{tabular}{lcccccccccc}
%    \toprule
%%    \multicolumn{2}{c}{Part}                   \\
%%    \cmidrule(r){1-2}
%    Method & Naive & iCARL & LUCIR & SSIL & POD & AANet & AFC & DER & Exploit & Oracle \\
%    \midrule
%%    Final CKA & 0.380 & 0.517 & 0.772 & 0.854 & 0.946 & 0.990 & 0.952 & 1.000 & 1.000 & - \\
%%    Mean CKA & 0.873 & 0.884 & 0.957 & 0.926 & 0.926 & 0.894 & 0.983 & 1.000 & 1.000 & - \\
%    \acc{\model'_0}{\data} & 61.0 & 61.8 & 62.1 & 62.5 & 62.9 & 62.5 & 62.9 & 60.0 & 62.9 & 60.4 \\
%    \acc{\model'_5}{\data} & 52.2 & 53.9 & 62.8 & 63.5 & 62.6 & 62.8 & 62.9 & 66.9 & 62.9 & 70.8 \\
%    DE (Eq.(~\ref{eq:data_efficiency})) & -0.85 & -0.76 & 0.07 & 0.10 & -0.03 & 0.03 & 0.00 & 0.66 & 0.00 & 1.00 \\
%%    Lack of: & Stab. & Stab. & Plast. & Plast. & Plast. & Plast. & Plast. & Scal. & Plast. & - \\
%    \bottomrule
%  \end{tabular}
%  }
%%  \vspace{2mm}
%\end{table}

%% file: tables/cil_benchmarks.tex
% !TEX root = ./../main.tex

%%%%%%%%%% Transposed

\begin{table*}
%  \vspace{-5mm}
  \caption{Average incremental accuracy and Final ImageNet accuracy of all compared CIL algorithms. $^\dagger$SSIL uses a different class ordering than all other methods; thus, inter-method comparisons with SSIL may not be appropriate. *Methods introduced in Section~\ref{sec:exploit}}
  \label{tab:cil}
  \vspace{-2mm}
  \centering
  \setlength{\tabcolsep}{5pt}
  \renewcommand{\arraystretch}{1.1}
  \scalebox{0.9}{
  \begin{tabular}{lccccccccccc}
    \toprule
%    \multicolumn{2}{c}{Part}                   \\
%    \cmidrule(r){1-2}
    Method & Naive & iCARL & LUCIR & SSIL$^\dagger$ & POD & AANet & AFC & DER & Exploit* & pDER* & Oracle \\
    \midrule
    Avg. Inc. Acc. & 41.4 & 31.5 & 55.6 & 65.8 & 66.3 & 57.1 & 66.4 & 69.1 & 67.2 & 69.7 & 72.8 \\
    \acc{\model_5}{\data} & 31.5 & 17.2 & 41.0 & 59.8 & 58.6 & 43.3 & 59.0 & 63.8 & 60.5 & 64.7 & 70.8\\
    \bottomrule
  \end{tabular}
  }
\vspace{-2mm}
\end{table*}

%% file: sections/conclusion.tex
% !TEX root = ./../main.tex

We took a deep dive into how effectively modern CIL algorithms address the stability-plasticity dilemma.
We introduced evaluation protocols that help us better understand the stability and plasticity of feature representations.
Our evaluations of recent works showed that many CIL methods are too fixated on the notion of alleviating catastrophic forgetting, to the extent that the feature extractor rarely learns any new concepts after the initial stage of training, namely on $\data_0$.
Based on this observation, we introduced two simple algorithms that improve upon an existing algorithm and exploit the shortcomings of the standard evaluation metrics for CIL research.
All in all, we hope that our findings will propel CIL research to focus more on stronger continual learning of feature representations.
%While the exact reasons of such fixation are unknown, we speculate that the limitations of commonly used CIL metrics play a major role.

%According Table~\ref{tab:cil}, LUCIR and AANet demonstrate weak performance, as their average incremental and final ImageNet accuracies are roughly 10\%p and 20\%p lower than POD, AFC, DER, SSIL, and Exploit.
%However, according to Table~\ref{tab:summary}, LUCIR and AANet both demonstrate similar \acc{\model'_0}{\data} and \acc{\model'_5}{\data} scores to all the aforementioned CIL algorithms.
%This suggests that while the feature representations of all compared methods, excluding Naive, iCARL, and DER, all have similar levels of discriminativeness, this is not well conveyed in CIL works, which only report average incremental accuracy.
%In other words, CIL research only appears to be making significant progress in terms of flawed metrics, but in reality, is somewhat stagnated, especially in terms of feature representation learning.
%All in all, we propose that future works in CIL research must place more emphasis on better feature representation learning, and leave it up to future works to figure out how this may be achieved in a scalable manner.

\vspace{-3mm}
\paragraph{Acknowledgements}
%This work was partly supported by Samsung Electronics Co., Ltd., and by the NRF Korea grant [No. 2022R1A2C3012210, Knowledge Composition via Task-Distributed Federated Learning] and the IITP grants [No.2022-0-00959, (Part 2) Few-Shot Learning of Causal Inference in Vision and Language for Decision Making; 2021-0-02068, Artificial Intelligence Innovation Hub; 2021-0-01343, Artificial Intelligence Graduate School Program (Seoul National University)] funded by the Korean government (MSIT).
This work was partly supported by Samsung Electronics Co., Ltd., and by the NRF Korea grant [No. 2022R1A2C3012210] and the IITP grants [No.2022-0-00959; 2021-0-02068; 2021-0-01343] funded by the Korean government (MSIT).

\iffalse
\paragraph{Ethics statement}
Fairness in AI systems has attracted a lot of attention in recent years, and has been fueling progressive research in the field of debiasing.
A potential ethical concern of class-incremental learning is that the model is prone to learning biases for the most recent set of classes.
In fact, classifier bias is a widely known issue in the class-incremental setting (of the compared CIL algorithms in our paper, SSIL and LUCIR explicity aim to reduce classifier bias).
Since our work focuses on the feature representations, it does not help to address the potentially harmful aspects of classifier bias in CIL models.

\paragraph{Reproducibility statement}
The codes for baseline models reported in our paper are all open-sourced by authors of previous works.
Please refer to Appendix~\ref{sup:implementation} for more implementation details of each baseline model, as well as links to the open-source codes.
\fi

%% file: sections/supp_body.tex
% !TEX root = ./../main.tex

\section{More details on partial-DER}
\label{sec:pder}
\input{tables/partial_der}

\subsection{Scalability issues in DER}

As mentioned Section 5.1, DER offers strong CIL performance but suffers from scalability.
For example, in our ImageNet-B500 5step setting, the last stage model for DER maintains 6 full ResNet-18 models.
To make matters worse, in the ImageNet-B500 10step setting, DER maintains 11 full ResNet-18 models.
This affects inference as well, since an input must pass through all feature extractors before classification.
%While DER takes a step in the right direction, we must be concerned about its scalability.
%In the B500-10step setting, for example, the final DER model has 11 feature extractors, and adding more tasks makes the model even heavier.
Although the authors do propose a masking/pruning scheme to reduce the memory complexity, they do not provide an implementation in their official code despite the fact that masking/pruning is an integral part of their algorithm.
Naturally, we raise certain doubts on the reproducibility of DER with masking, and question whether the benefits of improved performance so overwhelmingly outweigh the loss of scalability.

\subsection{Partial-DER ablations}
The pDER method that we introduced in Section 5.1 aims to improve scalability of DER by only maintaining a subset of layers from all stages.
This specific instance of pDER is the pDER Layer 4 variant, where only the parameters of ResNet Layer 4 are replicated and trained at each incremental stage.
We also test with other variants:

\begin{itemize}
    \item pDER Layer 3: replicate and train ResNet Layers 3 and 4; fix all parameters upto Layer 3
    \item pDER Layer 2: replicate and train ResNet Layers 2, 3, and 4; fix all parameters upto Layer 2
\end{itemize}

The results of our ablations are presented in Table~\ref{tab:partial_der}.
Surprisingly, we find that as we apply DER on deeper layers, the performance actually \textit{increases} across all metrics: \acc{\model'_5}{\data} increases by 0.6\%p, the average incremental accuracy improves by 0.5\%p, and \acc{\model_5}{\data} improves by 0.6\%p from pDER Layer 2 to Layer 4.
This is achieved all while reducing the GMACs for a single-input forward pass.

\section{Feature representations of ImageNet B0-10step models}
\begin{figure}[h!]
    \centering
    \includegraphics[width=0.5\textwidth]{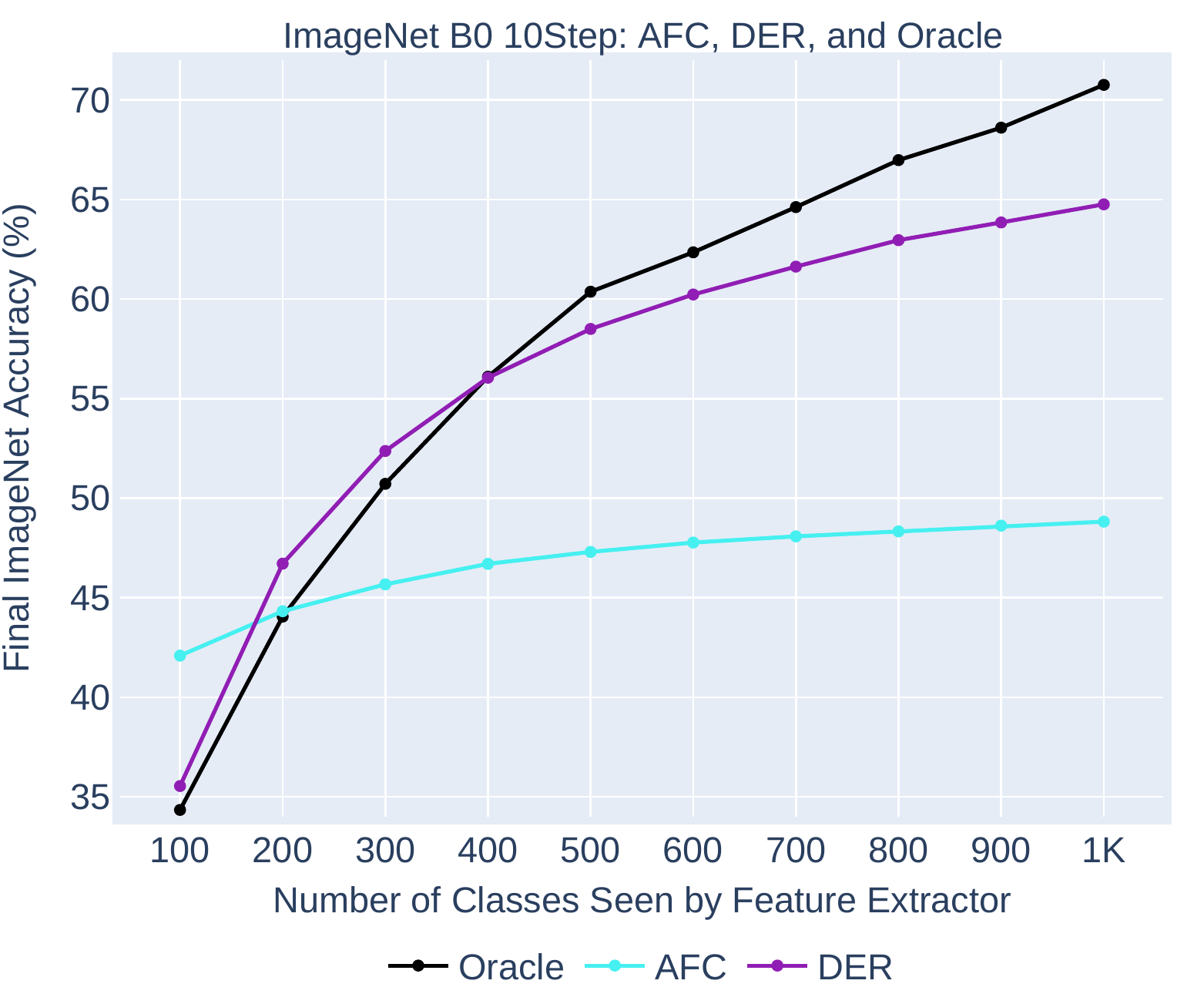}
    \caption{Accuracy on the ImageNet validation set after fine-tuning the classification layer of each incremental model (B0-10step setting) with the full ImageNet training data. The black line indicates the Oracle model trained on \{100, 200, ..., 1000\} classes, and serves as a point of reference for the performance on non-incremental settings.}
    \label{fig:b0_10step}
\end{figure}

In Figure~\ref{fig:b0_10step} we present the classifier finetuning analysis of DER, AFC, and Oracle models on the ImageNet B0-10step setting.
The purpose of this plot is to demonstrate that we observe a lack of plasticity for CIL models even in the B0-10step setting, which does not use a pre-trained base model.
The purple line shows the progression of finetuning accuracy as the DER model is trained continually on 100 classes at a time.
As seen in the plot, it closely follows the Oracle, but falls short by around 5\% points by the last stage.
On the other hand, the blue line, which represents AFC, shows a much different trend.
At 100 classes, AFC has a much higher (base) accuracy of 42.1\% (vs. $\sim 35\%$ of DER and Oracle), while the final accuracy is much lower at 48.8\% (vs. $70.8\%$ for the Oracle and $64.8\%$ for DER).
The 6.7\% increase in accuracy is rather trivial when considering that the final model has seen 9 times the data of the base model.
Thus, even in the B0-10step setting, we observe that AFC significantly lacks plasticity in its feature representations.

%\clearpage
\section{Mini-batch CKA}

Eq. (3) of our main paper describes CKA, which takes as input $\mathbf{X} \in \mathbb{R}^{b \times z_1}$ and $\mathbf{Y} \in \mathbb{R}^{b \times z_2}$.
In our case, $b$ is equivalent to the number of samples in $\data_0$, \textit{i.e.,} $|\data_0| = 25\text{K}$ for the ImageNet validation set.
Storing matrices $\mathbf{X} \in \mathbb{R}^{25000 \times z_1}$ and $\mathbf{Y} \in \mathbb{R}^{25000 \times z_2}$, where $z_1$ and $z_2$ are the dimensions of the flattened output features, requires excessive memory especially when CKA is computed on the GPU.
Furthermore, we require $\mathbf{X}$ and $\mathbf{Y}$ for all layers of ResNet-18, which makes storing such matrices even less feasible.

To alleviate the memory burden, we utilize the mini-batch variant of CKA proposed by~\cite{dodeepandwide}.
The main difference is that the mini-batch CKA uses an unbiased estimator of HSIC:
\begin{equation}
    \text{HSIC}_1 (\mathbf{K}, \mathbf{L}) = \frac{1}{n (n - 3)} \left( \text{tr}(\tilde{\mathbf{K}}\tilde{\mathbf{L}}) + \frac{\mathbf{1}^\mathsf{T} \mathbf{\tilde{\mathbf{K}}} \mathbf{1}\mathbf{1}^\mathsf{T} \mathbf{\tilde{\mathbf{L}}} \mathbf{1}}{(n - 1)(n - 2)} - \frac{2}{n - 2} \mathbf{1}^\mathsf{T} \mathbf{\tilde{\mathbf{K}}}\mathbf{\tilde{\mathbf{L}}} \mathbf{1}\right),
\end{equation}
where $\tilde{\mathbf{K}}$ and $\tilde{\mathbf{L}}$ are equivalent to $\mathbf{K}$ and $\mathbf{L}$ with their diagonals set to zero, and $n$ denotes the size of the mini-batch.
Thus, given activation matrices $\mathbf{X}_i \in \mathbb{R}^{n \times z_1}$ and $\mathbf{Y}_i \in \mathbb{R}^{n \times z_2}$, the mini-batch CKA formulates to:
\begin{equation}
    \text{CKA}_{\text{mini-batch}} = \frac{\sum_{i=1}^k \text{HSIC}_1 (\mathbf{X}_i \mathbf{X}_i^{\mathsf{T}}, \mathbf{Y}_i \mathbf{Y}_i^{\mathsf{T}})}{\sqrt{\sum_{i=1}^k \text{HSIC}_1 (\mathbf{X}_i \mathbf{X}_i^{\mathsf{T}}, \mathbf{X}_i \mathbf{X}_i^{\mathsf{T}})} \sqrt{\sum_{i=1}^k \text{HSIC}_1 (\mathbf{Y}_i \mathbf{Y}_i^{\mathsf{T}}, \mathbf{Y}_i \mathbf{Y}_i^{\mathsf{T}})}},
\end{equation}
where $k$ denotes the number of iterations.
Following~\cite{dodeepandwide}, we use a batch size of $n = 256$ and iterate over $\data_0$ 10 times to compute mini-batch CKA.

\section{More implementation details}
\subsection{Classifier types}
\label{sub:classifier}
\paragraph{Linear classifier}
The linear classifier is a simple matrix multiplication with an added bias term, and is formulated as below:
\begin{equation}
    \mathbf{y} = \mathbf{W}\mathbf{x} + \mathbf{b},
\end{equation}
where $\mathbf{y} \in \mathbb{R}^{c \times 1}$ denotes the output logit vector (with $c$ classes), $\mathbf{x} \in \mathbb{R}^{z \times 1}$ denotes the $z$-dimensional output of the feature extractor, $\mathbf{b} \in \mathbb{R}^{c \times 1}$ denotes the bias term, and $\mathbf{W} \in \mathbb{R}^{c \times z}$ denotes the weight matrix.
Note that some implementations do not use the bias term.

\paragraph{Cosine classifier}
The output of the cosine classifier is formulated as:
\begin{equation}
    \mathbf{y}_i = s \cdot \frac{\mathbf{W}_i\mathbf{x}}{\lVert \mathbf{W}_i\rVert  \lVert\mathbf{x}\rVert},
\end{equation}
where $\mathbf{y}_i$ denotes the logit for the $i$-th class, and $\mathbf{W}_i \in \mathbb{R}^{1 \times z}$ denotes the $i$-th row vector of the weight matrix, $\mathbf{W}$.
Finally, $s$ denotes the scale factor, which may be either fixed or set as a learnable parameter.
Note that the scale factor is used solely for training purposes, but does not affect the prediction output during evaluation.

\subsection{CIL algorithm implementation details}
\label{sup:implementation}
\paragraph{Common details}
All compared CIL models use a ResNet-18 backbone with varying classifier types.
When the cosine classifier is used, the output of the feature extractor is not subject to ReLU activation, following POD~\cite{douillard2020podnet}.
Furthermore, all implementations (except SSIL) employ the same class orderings for ImageNet.

\paragraph{Naive and Oracle}
For the Naive and Oracle models, we employ the cosine classifier with a fixed scale factor of $s = 24$.
We use a batch size of 512, an initial learning rate of $\text{lr} = 0.1$ and a polynomial learning rate decay scheme with a power of 0.9 over the course of 120 epochs.
For data augmentation, we use the standard sequence: \texttt{\{random resized crop, random horizontal flip\}}.
For the naive model, instead of the herding selection scheme, we randomly select 20 exemplars from each class to fill up the exemplar set.

\paragraph{POD and AFC}
For POD and AFC, we train models using their official codes\footnote[1]{POD: \url{https://github.com/arthurdouillard/incremental_learning.pytorch}} \footnote[2]{AFC: \url{https://github.com/kminsoo/AFC}}.
One important detail to note is that both POD and AFC originally used a modified version of ResNet, where the first convolution layer (\texttt{conv1}) had \texttt{kernel\_size=3, stride=1, padding=1} as opposed to \texttt{kernel\_size=7, stride=2, padding=3} from the original ResNet.
Thus, we fixed this detail and re-ran their code to obtain the POD and AFC models.
While POD and AFC both use the local similarity classifier~\cite{douillard2020podnet, AFC}, we use the cosine classifier when retraining the classification layer.

\paragraph{iCARL, LUCIR, and AANet}
For iCARL, LUCIR, and AANet, we obtained the trained models from the Energy-based Latent Aligner (ELI)~\cite{ELI} codebase\footnote[3]{ELI: \url{https://github.com/JosephKJ/ELI}}.
Unfortunately, we do not evaluate ELI itself, since it requires the high-level distinction between previous and new classes for prediction.
iCARL, LUCIR, and AANet all use the cosine classifier, where the scale factor $s$ is also set as a trainable parameter.

\paragraph{SSIL}
We received the trained SSIL models directly from the authors.
SSIL models use an oridinary linear classifier with a bias term.
Note that the trained SSIL models used a different class ordering compared to all other methods, although this does not affect the conclusions made in our paper.

\paragraph{DER}
We use the official DER code\footnote[4]{DER: \url{https://github.com/Rhyssiyan/DER-ClassIL.pytorch}} to train DER models.
While their code does not provide configurations for ImageNet-1K experiments, we reproduced results using the details from the paper.
However, the authors do not provide code for the masking operation, and thus, the reported results are the ones for full DER without masking/pruning.
This makes the final DER models (for both the 5-step and 10-step settings) extremely large.
DER uses the ordinary linear classifier without the bias term.

\paragraph{Retraining the classifier on full ImageNet data}
To retrain the classifiers on full ImageNet data, we freeze all layers prior to the final classification layer, including Batch Normalization layers (whose running means and variances are fixed).
We select the appropriate classifier layer type depending on what type of classifier was used to train the original models.
We train the classifiers for 60 epochs, using the same batch size, learning rate, and learning rate decay schemes as detailed in the ''Naive and oracle'' paragraph.

\section{Changes in Task Similarity}
%%%%%%%%%%%%%
\begin{table}[h]
  \caption{Retrained classifier accuracy on the ImageNet-C~\cite{hendrycks2019robustness} B500-5step setting.}
  \label{tab:corrupted_imagenet}
  \centering
  \setlength{\tabcolsep}{3pt}
  \scalebox{0.9}{
  \begin{tabular}{lcccccc}
    \toprule
    Method & \acc{\model'_0}{\data} & \acc{\model'_1}{\data} & \acc{\model'_2}{\data} & \acc{\model'_3}{\data} & \acc{\model'_4}{\data} & \acc{\model'_5}{\data}\\
     & Clean & Gaussian Noise & Contrast Shift & Pixelate & Impulse & Blur \\
    \midrule
    AFC & 70.8 & 70.8 & 70.9 & 71.6 & 71.7 & 72.1 \\
    DER & 67.3 & 70.3 & 72.3 & 76.2 & 78.3 & 79.3 \\
    \bottomrule
  \end{tabular}
  }
  \vspace{-4mm}
\end{table}

To investigate whether our analyses are consistent when the task similarity changes, we design an experiment under the CIL setting by injecting larger distribution shifts in each sequential task.
More specifically, for each $\data_{i>0}$ in the ImageNet B500-5step setting, we apply perturbations/corruptions~\cite{hendrycks2019robustness} to the input images instead of training with the original images.
We select a unique type of perturbation for each incremental stage and apply this perturbation for all samples within a given stage, \eg all samples of $\data_{1}$ are perturbed by random gaussian noise, while samples of $\data_{2}$ are perturbed by contrast shift, and so on.
By doing so, we inject a domain shift between samples of different stages, thereby decreasing the task similarity.
We train AFC and DER under this setting and report the results of our analysis in Table~\ref{tab:corrupted_imagenet}.
Table~\ref{tab:corrupted_imagenet} shows that the trend is clear; while DER shows significant improvement from $\model_0$ to $\model_5$ (\ie $\Delta\model_5 = 12.0$), AFC exhibits relatively minor improvements in performance (\ie $\Delta\model_5 = 1.3$).
Thus, our findings are consistent even when task similarity changes.

\section{AANet CKA Anomaly}
\label{sec:aanet_cka}
The CKA curve for AANet is quite interesting since it exhibits some erratic behavior.
We observe 4 major drops, which start at layers 8, 20, 32, and 44.
Interestingly, these layers are all 12 layers apart, and all correspond to a specific \texttt{3x3} convolution within the ResNet-18 architecture.
We believe that this is due to the unique design of AANet, where two models (one “stable” and one “plastic”) are fused together after each ResNet Layer.
Our hypothesis is that the representations diverge between the AANet fusion steps, and converge again when features of both branches are added together in the fusion step.

\section{GDumb}
\label{sec:gdumb}
\begin{table}[h]
  \caption{Retrained classifier performance for GDumb (with CutMix~\cite{yun2019cutmix}).}
  \label{tab:gdumb}
  \centering
  \setlength{\tabcolsep}{7pt}
  \renewcommand{\arraystretch}{1.1}
  \scalebox{1.0}{
  \begin{tabular}{cccccc}
    \toprule
%    \multicolumn{2}{c}{Part}                   \\
%    \cmidrule(r){1-2}
     \acc{\model'_0}{\data} & \acc{\model'_1}{\data} & \acc{\model'_2}{\data} & \acc{\model'_3}{\data} & \acc{\model'_4}{\data} & \acc{\model'_5}{\data} \\
    \midrule
    61.0 & 56.4 & 56.5 & 57.0 & 56.8 & 56.9 \\
    \bottomrule
  \end{tabular}
  }
%  \vspace{2mm}
\end{table}

\paragraph{Results}
Much like our Exploit, the motivation behind GDumb is to question the general progress in continual learning research.
Thus, we do not expect the GDumb model to exhibit favorable properties, and the results in Table~\ref{tab:gdumb} match our expectations; GDumb starts with a base model (\acc{\model'_0}{\data} = 61.0), but the becomes less performant from the first incremental stage ($56.4 \leq$ \acc{\model'_i}{\data} $\leq 56.9, \forall i > 0$).

\paragraph{Implementation} Since GDumb does not experiment on ImageNet-1K, we reproduce GDumb in our own code base.
One important detail is that for any setting with pre-trained models (e.g., ImageNet B500-5step), GDumb will use the model pre-trained on 500 classes as the initialization for each incremental step.
Although their paper states that a model is trained “from scratch”, the official GDumb code actually re-loads the pre-trained model for each incremental step (for methods that require pre-training).
Also, following the official implementation, we include CutMix~\cite{yun2019cutmix} for GDumb, which is not used for any of the other compared methods.

\section{Full 5 step subset accuracies}
\begin{figure}[t]
    \centering
    \includegraphics[width=0.9\textwidth]{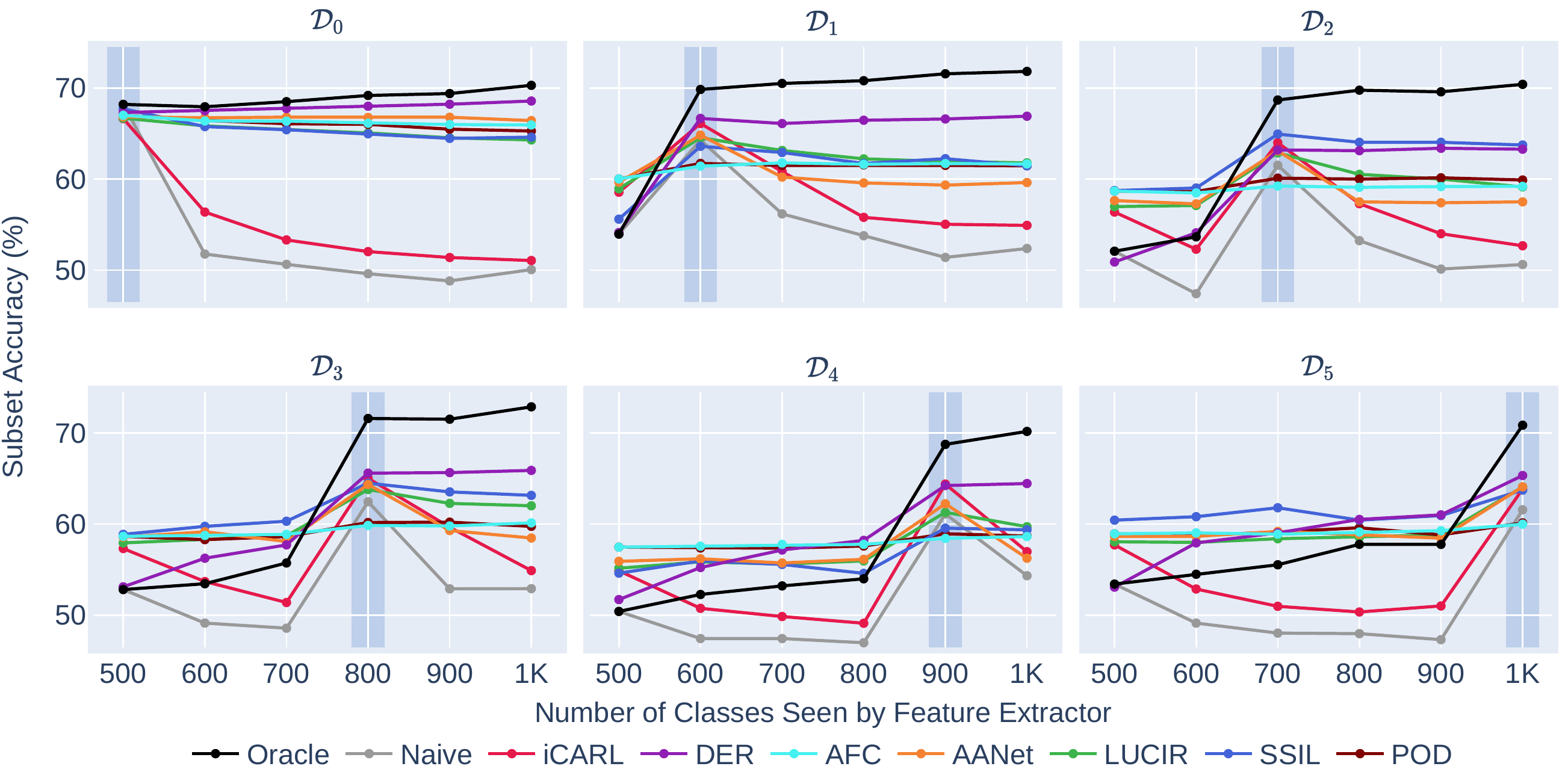}
    \vspace{-2mm}
    \caption{B500-10step subset accuracies for all methods. Note that SSIL uses a different class ordering.
    We highlight the region for model $\model'_j$ in plot $\classes_i$, where $i = j$.}
    \label{fig:5step_subset_all}
\end{figure}

We present the full ImageNet B500-5step subset accuracies in Figure~\ref{fig:5step_subset_all}.
\label{sec:5step_subset_full}

\begin{figure}[t]
    \centering
    \includegraphics[width=0.9\textwidth]{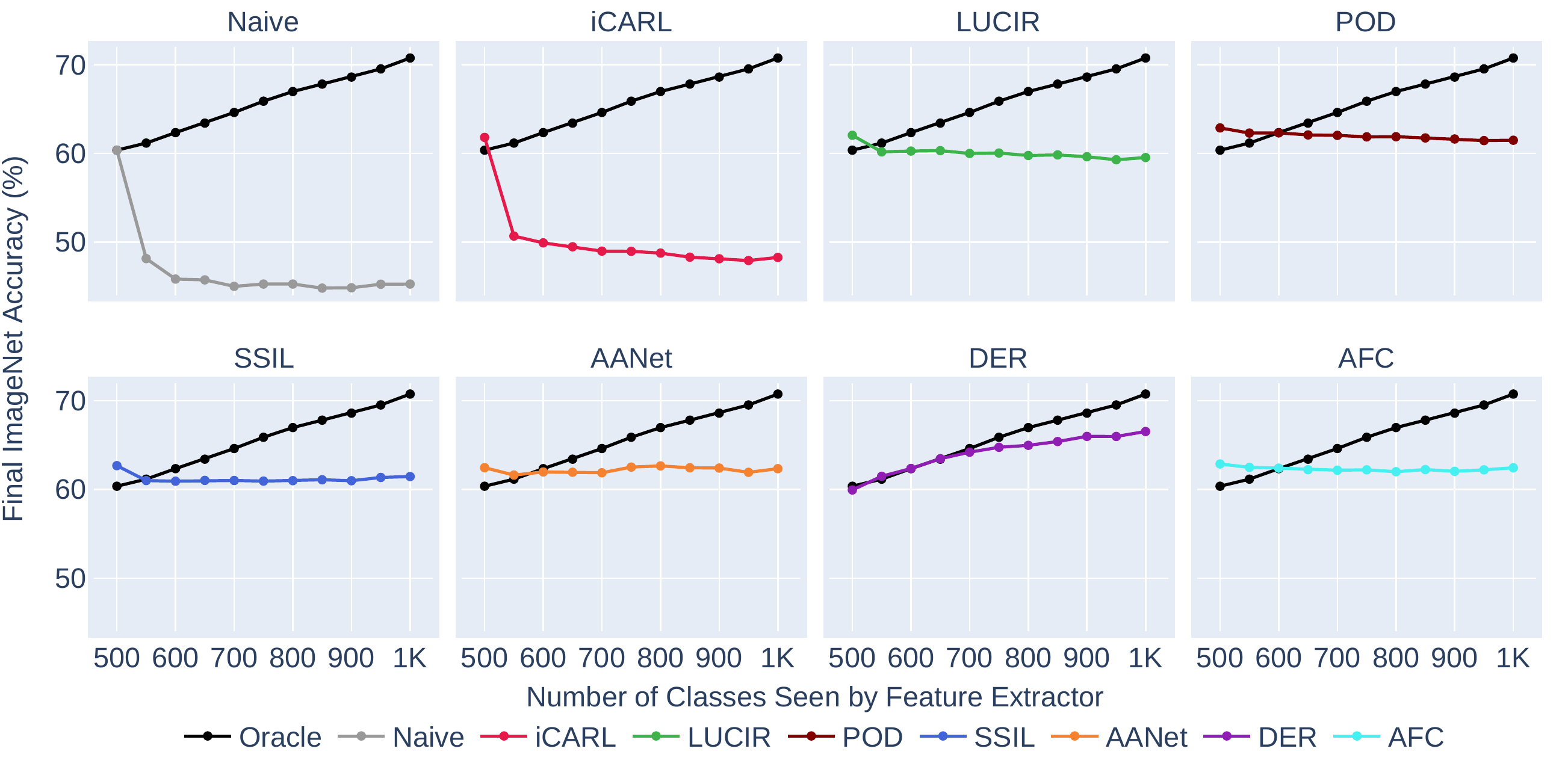}
    \vspace{-2mm}
    \caption{Accuracy on the ImageNet validation set after fine-tuning the classification layer of each incremental model (B500-10step setting) with the full ImageNet training data. The black line indicates an oracle model trained on \{500, 550, 600, ..., 1000\} classes, and serves as a point of reference for the performance on non-incremental settings.}
    \label{fig:10step_total_accs}
\end{figure}

\section{10 step results}
\label{sec:b500-10step}
In this section, we present figures for analyses on the ImageNet-1K B500-10step setting.
Overall, the evaluated CIL algorithms all show similar trends in both the B500-5step and B500-10step settings.
Thus, the B500-10step results serve to validate our observations made on the B500-5step setting.

%For ease of viewing, we also include HTML files for each figure in the main paper and appendix.
%These HTML files are interactive, meaning that viewers can inspect individual points, filter plots by algorithm (by clicking or double-clicking the corresponding label in the legend), and rescale the axis as needed.

\begin{figure}[t]
    \centering
    \includegraphics[width=0.9\textwidth]{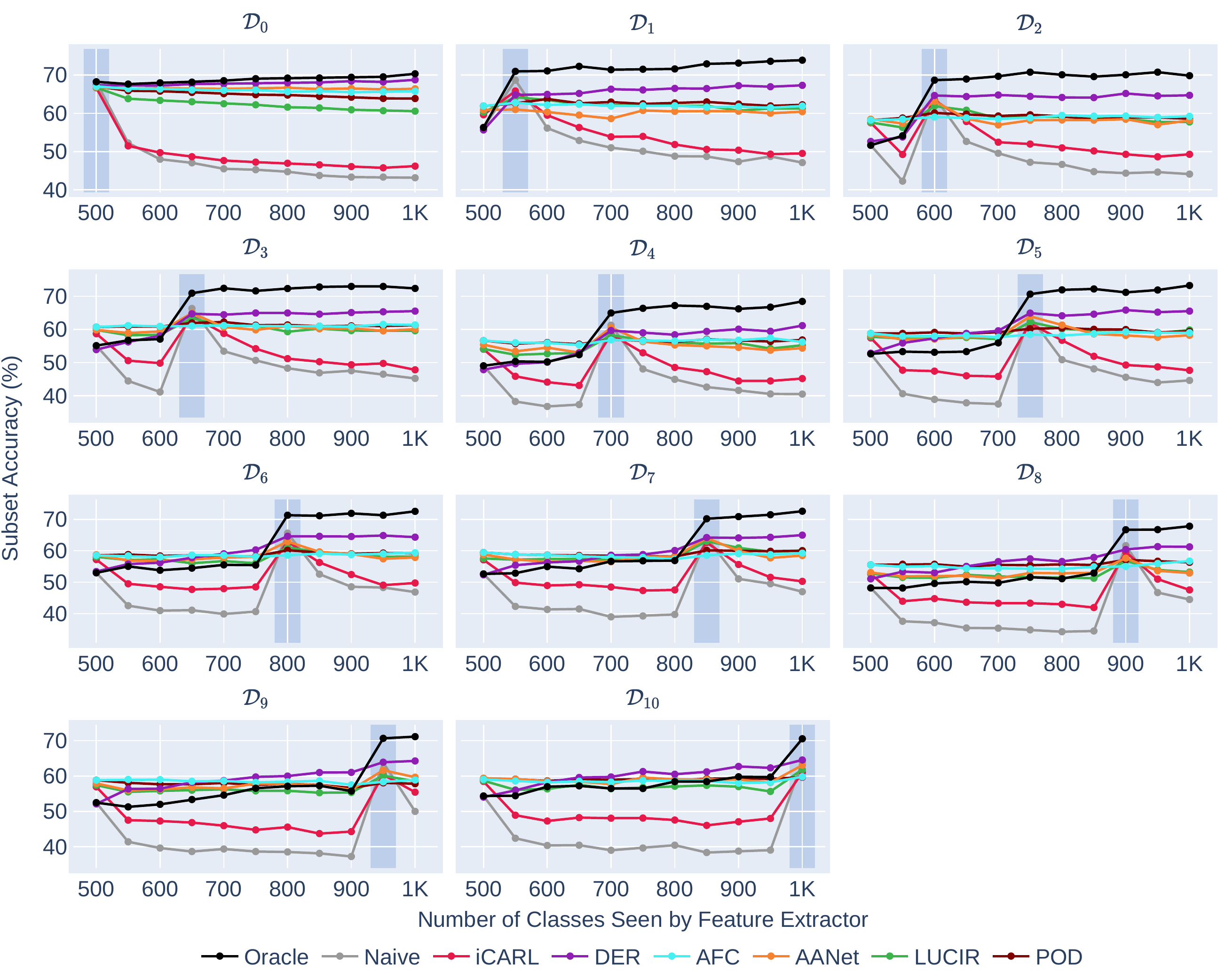}
    \vspace{-2mm}
    \caption{B500-10step subset accuracies for all methods. Note that SSIL uses a different class ordering.
    We highlight the region for model $\model'_j$ in plot $\classes_i$, where $i = j$.}
    \label{fig:10step_subset_all}
\end{figure}
\begin{figure}[h]
    \centering
    \includegraphics[width=0.9\textwidth]{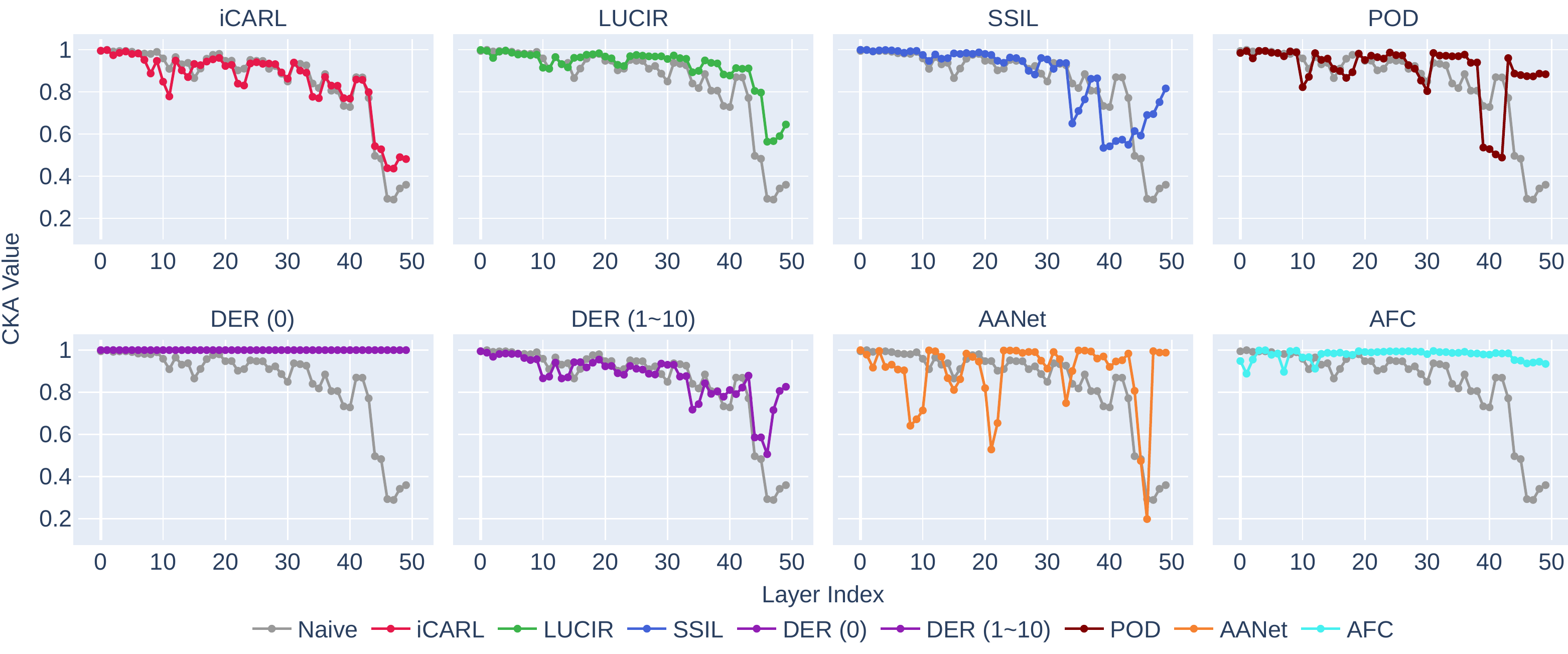}
    \vspace{-2mm}
    \caption{Same-layer CKA values between $\model_0$ and $\model_{10}$ for incremental models trained with each CIL algorithm on the ImageNet B500 10-step setting. The x-axis spans the layer index of ResNet-18, while the y-axis represents CKA. CKA is evaluated using the $\data_0$ validation set. Each plot is accompanied by the CKA for the naive model, which acts as a point of reference.}
    \label{fig:cka_10step}
\end{figure}

%% file: tables/partial_der.tex
% !TEX root = ./../main.tex

%%%%%%%%%% Transposed
\begin{table}[h]
  \caption{Performance of Partial-DER, with varying branch locations.}
  \label{tab:partial_der}
  \centering
  \setlength{\tabcolsep}{7pt}
  \renewcommand{\arraystretch}{1.1}
  \scalebox{0.9}{
  \begin{tabular}{lcccc}
    \toprule
%    \multicolumn{2}{c}{Part}                   \\
%    \cmidrule(r){1-2}
    Method & DER & pDER (Layer 2) & pDER (Layer 3) & pDER (Layer 4) \\
    \midrule
    \acc{\model'_0}{\data} & 60.0 & 60.0 & 60.0 & 60.0 \\
    \acc{\model'_5}{\data} & 66.9 & 67.8 & 68.0 & \textbf{68.4} \\
%    DE (Eq.(~\ref{eq:data_efficiency})) & 0.66 & 0.75 & 0.77 & 0.81 \\
    $\Delta\model'_5$ & 6.9 & 7.8 & 8.0 & \textbf{8.4} \\
    Avg. Inc. Acc. & 69.1 & 69.2 & 69.4 & \textbf{69.7} \\
    \acc{\model_5}{\data} & 63.8 & 64.1 & 64.4 & \textbf{64.7} \\
    GMACs ($\fe_5$) & 10.9 & 8.0 & 5.9 & \textbf{3.9} \\
    \bottomrule
  \end{tabular}
  }
%  \vspace{2mm}
\end{table}

%\acc{\model'_0}{\data} & 41.4 & 31.5 & 55.6 & 65.8 & 66.3 & 57.1 & 66.4 & 69.1 & 67.2 & 72.8 \\

%% file: main.bbl
\begin{thebibliography}{10}\itemsep=-1pt

\bibitem{SSIL}
Hongjoon Ahn, Jihwan Kwak, Subin Lim, Hyeonsu Bang, Hyojun Kim, and Taesup
  Moon.
\newblock {SS-IL:} separated softmax for incremental learning.
\newblock In {\em ICCV}, 2021.

\bibitem{Ben2010}
Shai Ben-David, John Blitzer, Koby Crammer, Alex Kulesza, Fernando Pereira, and
  Jennifer~Wortman Vaughan.
\newblock A theory of learning from different domains.
\newblock {\em Mach. Learn.}, 2010.

\bibitem{cortes_cka}
Corinna Cortes, Mehryar Mohri, and Afshin Rostamizadeh.
\newblock Algorithms for learning kernels based on centered alignment.
\newblock {\em JMLR}, 2012.

\bibitem{NTKOverlap2021}
T. Doan et~al.
\newblock A theoretical analysis of catastrophic forgetting through the ntk
  overlap matrix.
\newblock In {\em AISTATS}, 2021.

\bibitem{vit}
Alexey Dosovitskiy, Lucas Beyer, Alexander Kolesnikov, Dirk Weissenborn,
  Xiaohua Zhai, Thomas Unterthiner, Mostafa Dehghani, Matthias Minderer, Georg
  Heigold, Sylvain Gelly, Jakob Uszkoreit, and Neil Houlsby.
\newblock An image is worth 16x16 words: Transformers for image recognition at
  scale.
\newblock {\em ICLR}, 2021.

\bibitem{douillard2020podnet}
Arthur Douillard, Matthieu Cord, Charles Ollion, and Thomas Robert.
\newblock {PODNet: Pooled Outputs Distillation for Small-Tasks Incremental
  Learning}.
\newblock In {\em ECCV}, 2020.

\bibitem{Ganin2015}
Yaroslav Ganin and Victor Lempitsky.
\newblock Unsupervised domain adaptation by backpropagation.
\newblock In {\em ICML}, 2015.

\bibitem{HSIC}
Arthur Gretton, Olivier Bousquet, Alex Smola, and Bernhard Sch{\"o}lkopf.
\newblock Measuring statistical dependence with hilbert-schmidt norms.
\newblock In {\em Algorithmic Learning Theory}, 2005.

\bibitem{he2016deep}
Kaiming He, Xiangyu Zhang, Shaoqing Ren, and Jian Sun.
\newblock {Deep Residual Learning for Image Recognition}.
\newblock In {\em CVPR}, 2016.

\bibitem{hendrycks2019robustness}
D. Hendrycks and T. Dietterich.
\newblock Benchmarking neural network robustness to common corruptions and
  perturbations.
\newblock In {\em ICLR}, 2019.

\bibitem{hou2019learning}
Saihui Hou, Xinyu Pan, Chen~Change Loy, Zilei Wang, and Dahua Lin.
\newblock {Learning a Unified Classifier Incrementally via Rebalancing}.
\newblock In {\em CVPR}, 2019.

\bibitem{ioffe2015batch}
Sergey Ioffe and Christian Szegedy.
\newblock {Batch Normalization: Accelerating Deep Network Training by Reducing
  Internal Covariate Shift}.
\newblock In {\em ICML}, 2015.

\bibitem{ELI}
KJ Joseph, Salman Khan, Fahad~Shahbaz Khan, Rao~Muhammad Anwar, and Vineeth
  Balasubramanian.
\newblock Energy-based latent aligner for incremental learning.
\newblock In {\em CVPR}, 2022.

\bibitem{AFC}
Minsoo Kang, Jaeyoo Park, and Bohyung Han.
\newblock Class-incremental learning by knowledge distillation with adaptive
  feature consolidation.
\newblock In {\em CVPR}, 2022.

\bibitem{kirkpatrick2017overcoming}
James Kirkpatrick, Razvan Pascanu, Neil Rabinowitz, Joel Veness, Guillaume
  Desjardins, Andrei~A Rusu, Kieran Milan, John Quan, Tiago Ramalho, Agnieszka
  Grabska-Barwinska, et~al.
\newblock {Overcoming Catastrophic Forgetting in Neural Networks}.
\newblock {\em Proceedings of the national academy of sciences}, 2017.

\bibitem{cka_kornblith}
Simon Kornblith, Mohammad Norouzi, Honglak Lee, and Geoffrey Hinton.
\newblock Similarity of neural network representations revisited.
\newblock In {\em ICML}, 2019.

\bibitem{dta_iccv2019}
Seungmin Lee, Dongwan Kim, Namil Kim, and Seong-Gyun Jeong.
\newblock Drop to adapt: Learning discriminative features for unsupervised
  domain adaptation.
\newblock In {\em ICCV}, 2019.

\bibitem{liu2021adaptive}
Yaoyao Liu, Bernt Schiele, and Qianru Sun.
\newblock {Adaptive Aggregation Networks for Class-Incremental Learning}.
\newblock In {\em CVPR}, 2021.

\bibitem{nam2016learning}
Hyeonseob Nam and Bohyung Han.
\newblock Learning multi-domain convolutional neural networks for visual
  tracking.
\newblock In {\em CVPR}, 2016.

\bibitem{dodeepandwide}
Thao Nguyen, Maithra Raghu, and Simon Kornblith.
\newblock Do wide and deep networks learn the same things? uncovering how
  neural network representations vary with width and depth.
\newblock In {\em ICLR}, 2021.

\bibitem{noh2015learning}
Hyeonwoo Noh, Seunghoon Hong, and Bohyung Han.
\newblock {Learning Deconvolution Network for Semantic Segmentation}.
\newblock In {\em ICLR}, 2015.

\bibitem{park2019continual}
Dongmin Park, Seokil Hong, Bohyung Han, and Kyoung~Mu Lee.
\newblock Continual learning by asymmetric loss approximation with single-side
  overestimation.
\newblock In {\em ICCV}, 2019.

\bibitem{park2021class}
Jaeyoo Park, Minsoo Kang, and Bohyung Han.
\newblock Class-incremental learning for action recognition in videos.
\newblock In {\em ICCV}, 2021.

\bibitem{prabhu2020gdumb}
Ameya Prabhu, Philip~HS Torr, and Puneet~K Dokania.
\newblock {GDumb: A Simple Approach that Questions Our Progress in Continual
  Learning}.
\newblock In {\em ECCV}, 2020.

\bibitem{clip}
Alec Radford, Jong~Wook Kim, Chris Hallacy, Aditya Ramesh, Gabriel Goh,
  Sandhini Agarwal, Girish Sastry, Amanda Askell, Pamela Mishkin, Jack Clark,
  Gretchen Krueger, and Ilya Sutskever.
\newblock Learning transferable visual models from natural language
  supervision.
\newblock In {\em ICML}, 2021.

\bibitem{CLSequentialGame21}
K. Raghavan and P. Balaprakash.
\newblock Formalizing the generalization-forgetting trade-off in continual
  learning.
\newblock In {\em NeurIPS}, 2021.

\bibitem{CKADoVisionTransformers}
Maithra Raghu, Thomas Unterthiner, Simon Kornblith, Chiyuan Zhang, and Alexey
  Dosovitskiy.
\newblock Do vision transformers see like convolutional neural networks?
\newblock In {\em NeurIPS}, 2021.

\bibitem{rebuffi2017icarl}
Sylvestre-Alvise Rebuffi, Alexander Kolesnikov, Georg Sperl, and Christoph~H
  Lampert.
\newblock {iCaRL: Incremental Classifier and Representation Learning}.
\newblock In {\em CVPR}, 2017.

\bibitem{redmon2016you}
Joseph Redmon, Santosh Divvala, Ross Girshick, and Ali Farhadi.
\newblock {You Only Look Once: Unified, Real-Time Object Detection}.
\newblock In {\em CVPR}, 2016.

\bibitem{ILSVRC15}
Olga Russakovsky, Jia Deng, Hao Su, Jonathan Krause, Sanjeev Satheesh, Sean Ma,
  Zhiheng Huang, Andrej Karpathy, Aditya Khosla, Michael Bernstein,
  Alexander~C. Berg, and Li Fei-Fei.
\newblock {ImageNet Large Scale Visual Recognition Challenge}.
\newblock {\em IJCV}, 2015.

\bibitem{tsne}
Laurens van~der Maaten and Geoffrey Hinton.
\newblock Visualizing data using t-sne.
\newblock {\em Journal of Machine Learning Research}, 2008.

\bibitem{vaswani2017attention}
Ashish Vaswani, Noam Shazeer, Niki Parmar, Jakob Uszkoreit, Llion Jones,
  Aidan~N Gomez, Lukasz Kaiser, and Illia Polosukhin.
\newblock {Attention is All you Need}.
\newblock In {\em NIPS}, 2017.

\bibitem{yan2021dynamically}
Shipeng Yan, Jiangwei Xie, and Xuming He.
\newblock {DER: Dynamically Expandable Representation for Class Incremental
  Learning}.
\newblock In {\em CVPR}, 2021.

\bibitem{yun2019cutmix}
Sangdoo Yun, Dongyoon Han, Seong~Joon Oh, Sanghyuk Chun, Junsuk Choe, and
  Youngjoon Yoo.
\newblock {Cutmix: Regularization strategy to train strong classifiers with
  localizable features}.
\newblock In {\em ICCV}, 2019.

\bibitem{Zeiler_visualizingNN}
Matthew~D. Zeiler and Rob Fergus.
\newblock Visualizing and understanding convolutional networks.
\newblock In {\em ECCV}, 2014.

\bibitem{zhao2020maintaining}
Bowen Zhao, Xi Xiao, Guojun Gan, Bin Zhang, and Shu-Tao Xia.
\newblock {Maintaining Discrimination and Fairness in Class incremental
  Learning}.
\newblock In {\em CVPR}, 2020.

\end{thebibliography}
